\documentclass[journal]{IEEEtran}
\usepackage{amssymb}
\usepackage{graphicx, amssymb, amsmath, booktabs, url, graphics}

\begin{document}
%
\title{Feature Sensitive Label Fusion with Random Walker for Atlas-based Image Segmentation}
%
%
%

\author{Siqi Bao and Albert C. S. Chung
\thanks{This work was supported in part by the Hong Kong Research Grants Council under Grant 16203115.}
\thanks{S. Bao is with the Lo Kwee-Seong Medical Image Analysis Laboratory, Department of Computer Science and Engineering, The Hong Kong University of Science and Technology, Clear Water Bay, Hong Kong (e-mail: sbao@cse.ust.hk).}
\thanks{A. Chung is with the Lo Kwee-Seong Medical Image Analysis Laboratory, Department of Computer Science and Engineering, The Hong Kong University of Science and Technology, Clear Water Bay, Hong Kong (e-mail: achung@cse.ust.hk).}}

\maketitle

\begin{abstract}
In this paper, a novel label fusion method is proposed for brain magnetic resonance image segmentation. This label fusion method is formulated on a graph, which embraces both label priors from atlases and anatomical priors from target image. To represent a pixel in a comprehensive way, three kinds of feature vectors are generated, including intensity, gradient and structural signature. To select candidate atlas nodes for fusion, rather than exact searching, randomized k-d tree with spatial constraint is introduced as an efficient approximation for high-dimensional feature matching. Feature Sensitive Label Prior (FSLP), which takes both the consistency and variety of different features into consideration, is proposed to gather atlas priors. As FSLP is a non-convex problem, one heuristic approach is further designed to solve it efficiently. Moreover, based on the anatomical knowledge, parts of the target pixels are also employed as graph seeds to assist the label fusion process and an iterative strategy is utilized to gradually update the label map. The comprehensive experiments carried out on two publicly available databases give results to demonstrate that the proposed method can obtain better segmentation quality. 
\end{abstract}


%
\IEEEpeerreviewmaketitle

\section{Introduction}
The human brain is a complex neural system composing many anatomical structures. To study the functional and structural properties of its subcortical regions, image segmentation is a critical step in quantitative brain image analysis and clinical diagnosis. However, segmenting subcortical structures is difficult because they are small and often exhibit large variations in shape. Moreover, some structural boundaries are subtle or even missing in images. Although manual annotation is a standard procedure for obtaining quality segmentation, it is time-consuming and can suffer from inter- and intra-observer inconsistencies. In recent years, researchers have been focusing on developing automatic atlas-based segmentation methods which can effectively incorporate expert prior knowledge about the relationships between local intensity profiles and tissue labels. And many softwares have become available for brain image segmentation, such as FreeSurfer \cite{fischl2012freesurfer}, BrainSuite \cite{shattuck2002brainsuite}, BrainVoyage \cite{goebel2006analysis}, BrainVisa \cite{geffroy2011brainvisa} and so on. 

\begin{figure}
	\centering
	\includegraphics[width=0.82\linewidth]{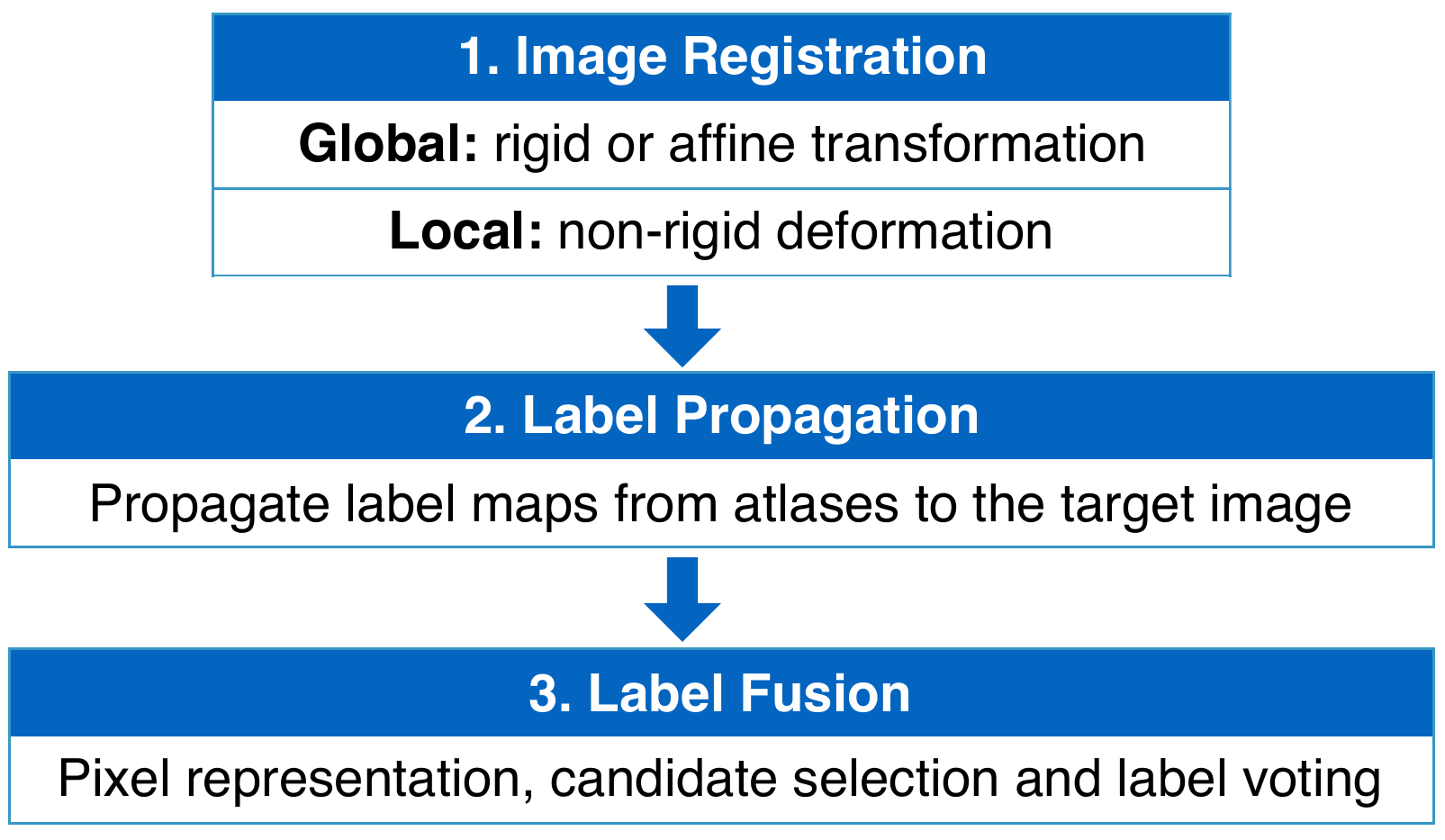}
	\caption{Overview of the main components in atlas-based segmentation.}
	\label{fig14}
\end{figure}

Atlas-based segmentation involves three main components, image registration between atlases and a target image, label propagation, and label fusion, as summarized in Fig. \ref{fig14}. To register images of intra-subject generated by different modalities, global transformation methods can be used, such as rigid or affine transformation. As for the registration of inter-subject or longitude analysis of intra-subject, global transformation is insufficient to estimate an accurate deformation field due to the high anatomical variabilities among these images. Local transformation, represented by non-rigid registration, has been proposed to deal with this problem. In non-rigid registration, the deformation field can be estimated using control points on the grid, with a combination of B-splines \cite{schnabel2001generic} or cosine basis functions \cite{ashburner1999nonlinear}.  To further improve the quality of anatomical or matching correspondences between two images, symmetric diffeomorphism \cite{avants2008symmetric} moves both images simultaneously along a geodesic path until meeting at the middle of normalization domain and then the whole path or deformation field can be obtained by uniting the two parts of geodesic paths. In the evaluation of 14 nonlinear deformation algorithms \cite{klein2009evaluation}, ANTs based on symmetric diffeomophism is selected as one of the best methods. While with a large number of target images to be labeled, the pairwise non-rigid registration methods can suffer from the expensive time consumption. 

\begin{figure*}
	\centering
	\includegraphics[width=0.92\linewidth]{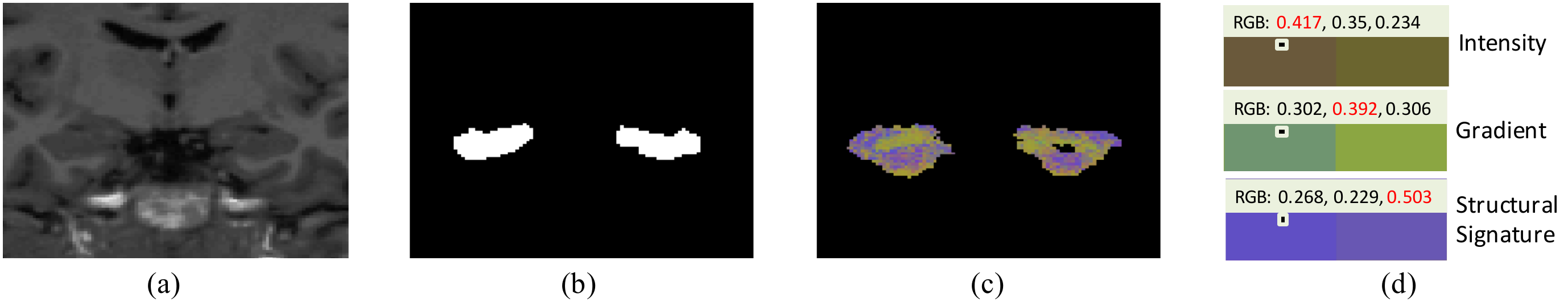}
	\caption{Effects of feature sensitivity. (a) Cropped target intensity image. (b) Manually labeled Hippocampus. (c) Color image to show the effects of feature sensitivity, with RGB values standing for the feature coefficients of intensity, gradient and structural signature respectively. (d) Three selected examples to illustrate their dominate features.}
	\label{fig11}
\end{figure*}
After image registration, the label maps can be propagated from atlases to the target image and multiple tissue labels can be collected for each image position, making label fusion a crucial final aggregation step for the reliable labeling of target images. A generative model for image segmentation based on label fusion is proposed in \cite{sabuncu2010generative} and different label fusion strategies are discussed. Majority voting is commonly used, while its accuracy can be adversely affected if the atlases are dissimilar. In voting using global weights, the similarity between each atlas and the target image is calculated and used as a weight during the label fusion process. Recently, more label fusion methods based on patches \cite{coupe2011patch,liao2013sparse,rousseau2011supervised,wang2013multi,tong2013segmentation,ta2014optimized} have been proposed, which were first introduced for image de-noising \cite{coupe2008optimized} and recently become more prevalent in medical image segmentation. 

Generally, there are three stages in patch-based label fusion. First, it is necessary to determine which kind of feature to be adopted as pixel representation. The conventional way is to collect pixel values inside the surrounding patch to formulate an intensity feature vector. To better reveal image changes, gradient magnitude is another commonly used feature information. However, it is not adequate enough to obtain quality segmentation if just relying on the above two kinds of features, as they can only capture local and low-level properties. Some advanced approaches have been proposed to extract high-level features to compensate for local limitations. In \cite{bai2015multi}, the contextual information, which estimates the relative relation between intensity values, is appended to form an augmented feature vector for cardiac image segmentation. 

With feature representation established, the second stage is to distinguish candidate pixels or patches for voting. In \cite{rousseau2011supervised}, to label a centre pixel in the target image, all surrounding small patches from atlases are utilized for weighted voting. To avoid the adverse effects from dissimilar patches, an extension has been proposed in \cite{coupe2011patch} which involves first ranking the small patches based on structure similarity, followed by combining the selected ones in the final labeling. Another patch selection method based on sparse representation was proposed by Liao et al. \cite{liao2013sparse}, which selects patch-based signatures with sparse logistic and the LASSO interface \cite{liu2009slep}. 

The third stage is to fuse the labels of candidate atlas nodes and the fusion strategies fall into two main categories: weighted voting and image patch reconstruction. It is a common way to first estimate the similarity between two patches by embedding their sum of squared difference to the Gaussian function and then to utilize the similarity value as the weight for voting \cite{coupe2011patch,rousseau2011supervised}. Besides the independent impact on target pixel, Joint Label Fusion \cite{wang2013multi} also takes the error correlation among atlas patches into consideration and tries to find the optimal weight for voting. For the second category, to reconstruct a target patch, the linear combination coefficients of atlas patches need to be optimized first and the label of the centre pixel can be then assigned to the class with a minimum reconstruction error \cite{tong2013segmentation}. 

Moreover, as shown in \cite{coupe2011patch,rousseau2011supervised}, the patch-based label fusion methods do not necessarily depend on the time-consuming non-rigid registration. While given the poor contrast condition in the brain Magnetic Resonance (MR) images and similar histogram profiles among adjacent structures, the label fusion with only affine transformation as processing becomes more challenging. As such, to compensate for the quality loss caused by the affine transformation, it raises the demand to design a more elegant label fusion process for brain MR image segmentation. Under the assumption that distinct features can assist the segmentation in a complementary way, in this paper, Feature Sensitive Label Prior (FSLP) is designed to capture label priors from atlases, whose process is distinct with the conventional label fusion at every stage.

As suggested in the segmentation of cardiac MR images, embracing more features besides intensity, such as contextual information, can help improve the segmentation quality \cite{bai2015multi}. For pixel representation, besides conventional intensity and gradient features, structural signature is introduced to extract the high-level property of each subcortical structure based on the Convolutional Neural Networks. During candidate node selection, rather than exact searching within a confined scope, the random k-d tree with a spatial constraint is put forward as an efficient approximation for high dimensional data matching. In the fusion stage, feature sensitivity is taken into account for the variance and consistency among various features. As FSLP is a non-convex problem, one heuristic method is further proposed to solve it by alternately dealing with two convex problems.

The \textbf{motivations} to introduce FSLP are two-fold. On the one hand, the contributions of distinct features are expected to be consistent during label fusion, i.e., they can reach an agreement when labeling a pixel. On the other hand, the impact of different features can change according to image conditions. For the flat regions away from structural boundaries, intensity and gradient are supposed to be more essential. As for the complex region near tissue bounders, structural signature should play a more significant role. The experimental result with our method also justifies the initial motivation, as shown in Fig. \ref{fig11}. The sub-figures (a) and (b) are a cropped target intensity image and its corresponding label map of the Hippocampus. In (c), for the pixels where atlases cannot make an agreement, the optimal feature coefficients estimated in FSLP are displayed as three channels of RGB. Three representative examples are selected to explain the dominant features in each pattern in (d). The color image in (c) demonstrates that the role of structural signature is more essential around tissue protrusions. For the other relatively flat regions, intensity and gradient matter more, which phenomenon also justifies our motivation to introduce feature sensitivity for label fusion. 

In addition to FSLP from atlases, anatomical priors from target image are also utilized to assist our graph-based label fusion process. Based on anatomical knowledge, to label a pixel which is deep inside or outside a subcortical structure is easier, while to label one which is located around the boundary is challenging. As such, rather than updating labels for all target pixels, those far away from structural border are selected as graph seeds and their influence can be propagated to other pixels through image lattice. Unlike the graph-based labeling constructed with both atlas and target nodes \cite{koch2015multi}, we further infer an equal but more concise graph to encode FSLP and anatomical prior, which only relies on target nodes. The objective energy function on the graph is formulated with Random Walker and can be solved as a discrete Dirichlet problem. To evaluate the proposed method, experiments have been carried out on two image databases and results demonstrate that our approach can obtain better performance as compared with other state-of-the-art methods.

Note that the preliminary version of this work has been published in the 19th International Conference on Medical Image Computing and Computer Assisted Intervention, MICCAI 2016. In this paper, 1) we extend our previous work by generating multiple features and introducing randomized k-d tree with spatial constraint for efficient high dimensional feature matching; 2) additional mathematical proofs, solutions together with illustrative examples are given in this work; 3) intensive experiments has been carried out to evaluate each component of our proposed method and comprehensive evaluations have been done with the state-of-the-art methods.

\section{Methodology}
In this paper, to obtain a more discriminative representation, three kinds of features are extracted and candidate nodes are selected for each pixel, which will be explained in Section \ref{FeaG} and Section \ref{FeaM}. Given the demands of consistency and variety among distinct feature vectors during label fusion, a novel method FSLP is proposed in Section \ref{FeaSLP} to deal with this dilemma, by collecting priors from atlases with feature sensitivity. Moreover, the pixels from target image are also selected based on anatomical knowledge, acting as anatomical prior. The whole label fusion process is modeled on an undirected graph and formulated under the framework of Random Walker, which is summarized in Section \ref{labelrw}.

\subsection{Feature Generation}\label{FeaG}
In medical images, the conventional feature utilized to represent a pixel is intensity values or gradient magnitudes in its surrounding cube. While these features are limited to local information and susceptible to adverse impacts from similar histogram profiles among tissues. As each subcortical structure in the brain has its own shape characteristics and structural properties, this kind of high-level features can be used to formulate a more discriminative representation. Recently it has been proven that the feature extraction ability of Convolutional Neural Networks (CNN) \cite{lecun1998gradient} has surpassed hand-crafted features, like SIFT \cite{lowe2004distinctive}, and CNN has brought significant improvements in image classification \cite{krizhevsky2012imagenet}, semantic segmentation \cite{long2015fully}, acoustic analysis \cite{sercu2015very} and so on. As such, in this paper, we propose to encode the high-level property for brain MR images with a feature vector extracted automatically using CNN.

\begin{figure}
\centering
\includegraphics[width=\linewidth]{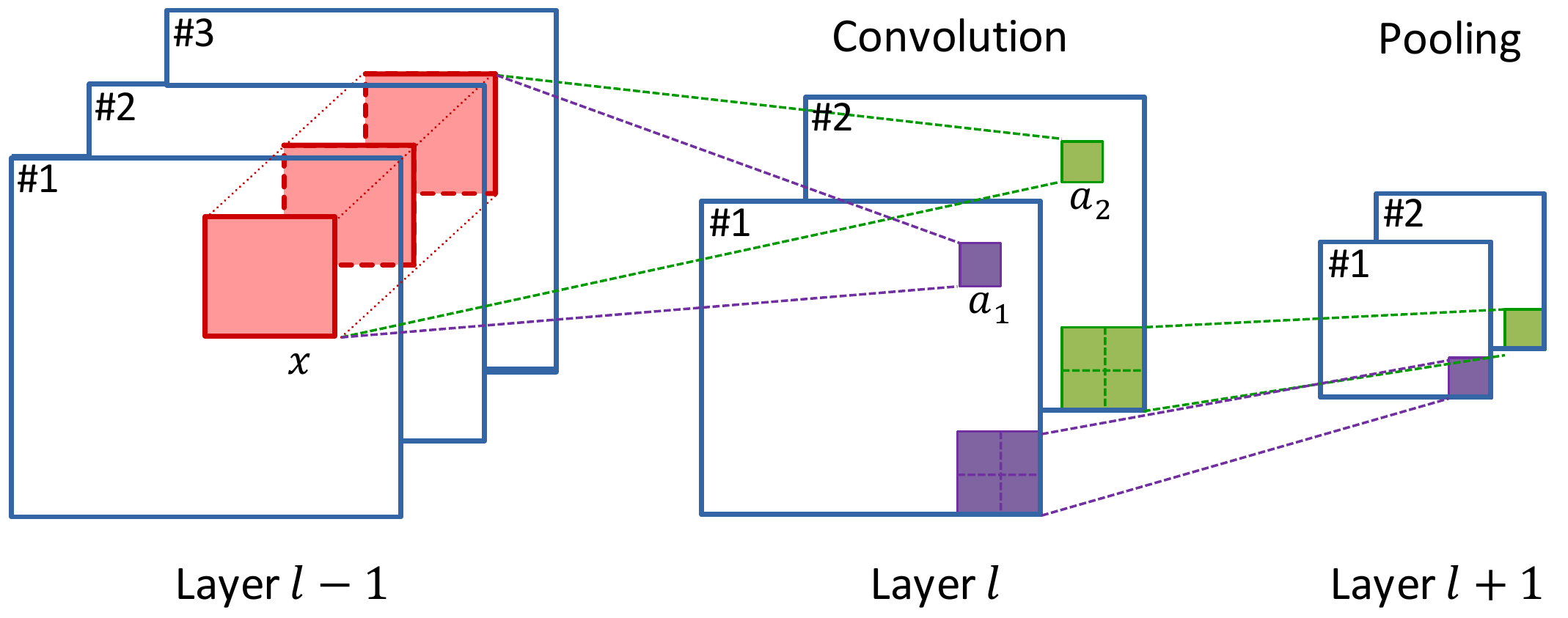}
\caption{Illustration of convolution and pooling layers. With three images from layer $l-1$ as input, two feature maps generated in layer $l$, each corresponding to one pair of $W$ and $b$, as stated in Equation (\ref{conv1}).}
\label{fig1}
\end{figure}

CNN is inspired from a biological visual mechanism, where neurons in the higher layer operate on a subregion of neurons in the lower layer. In CNN, there are two basic components: convolution and pooling layers, as illustrated in Fig. \ref{fig1}. To estimate the convolutional response $a_1$ or $a_2$ in layer $l$, pixels within the subregion of images in the last layer (Red Region, namely the receptive field) are chosen as input. The convolution step consists of linear operation and non-linear activation, which can be formulated as follows:
\begin{equation}\label{conv1}
a=f(Wx+b),
\end{equation}
where $a$ is the convolutional response, $f(\cdot)$ refers to the non-linear activation function, $x$ is the flattened input from the receptive field, $W$ is the weight vector and $b$ is the bias associated with the convolutional kernel. 

It is notable that each feature map in the convolution layer is assigned with a specific pair of $W$ and $b$. For the Purple pixel in the first feature map, its value can be estimated with $f(W_1x+b_1)$ and for the Green pixel in the second feature map, its value should be $f(W_2x+b_2)$. The number of feature maps in each layer can be preset during the design of the network architecture, while the parameters $W$ and $b$ need to be learned through training. 

For the non-linear activation $f(\cdot)$, the conventional way is to employ a sigmoid or tanh function. However, both can encounter the saturation problem and kill the gradients during backpropagation. Recently, non-saturated functions have become prevalent, such as Rectified Linear Unit (ReLU) \cite{nair2010rectified}, leaky ReLU \cite{maas2013rectifier} and some other variants. The experiment in \cite{krizhevsky2012imagenet} demonstrates that ReLU can accelerate the training speed up to 6 times faster than the tanh function. As such, in the paper, ReLU is chosen as the activation function and Equation \eqref{conv1} can then be rewritten as:
\begin{equation}\label{conv2}
a=\max(0,Wx+b).
\end{equation}

In CNN, the convolution and pooling layers are usually interwoven. The feature maps generated in the convolution layer can be regarded as input for the next pooling layer. As shown in Fig. \ref{fig1}, the patch in the $i$-th feature map of layer $l$ polls for the corresponding pixel in the $i$-th feature map of layer $l+1$. The pooling strategy used here can be either maximum or average pooling. From the example in Fig. \ref{fig1}, it can also be noticed that pooling can only shrink the feature maps, while leaving their amounts unchanged.

\begin{figure}
\centering
\includegraphics[width=\linewidth]{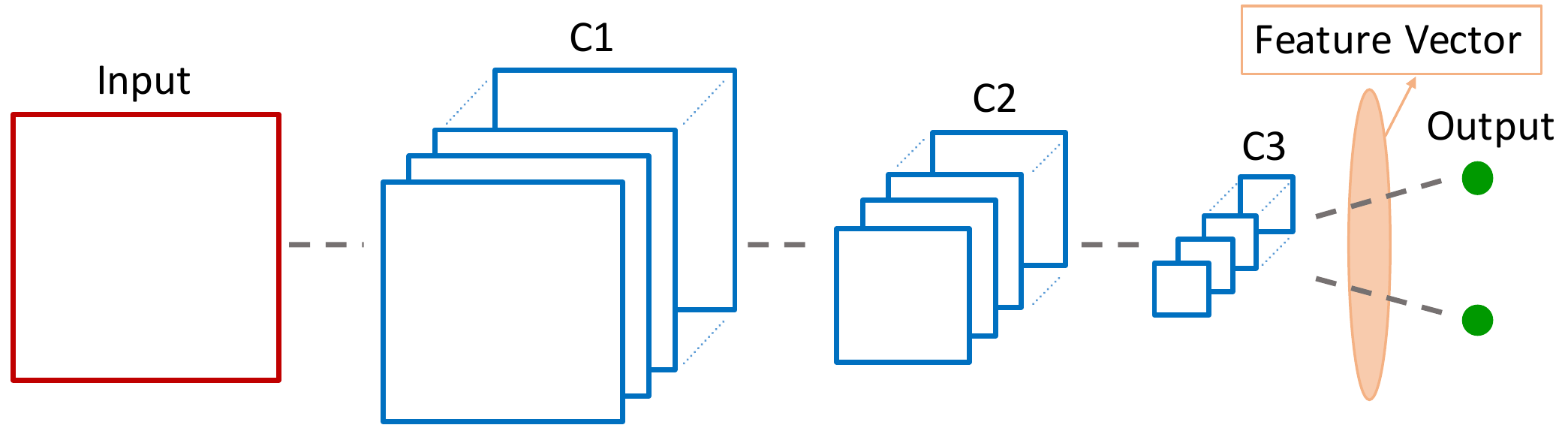}
\caption{CNN architecture. Red: 2D input patch; Blue: convolution layers; Green: output layer; Orange: feature vector to the output layer.}
\label{fig5}
\end{figure}

In this paper, to capture the high-level properties of subcortical structures, CNN is utilized to extract the structural signature from brain MR images. Fig. \ref{fig5} illustrates the architecture of the employed network. There are seven layers in the network, including six alternating convolution (C1, C2 and C3) and average pooling (omitted as dashed lines) layers, and one output layer. The input to the network is a 2D patch with the size of $20\times 20$ pixels and the two nodes in the output layer refer to the probability of each class. The feature vector to the output layer is extracted and regarded as the structural signature. Detailed parameter settings of the network can be found in Table \ref{tab1}. 

To train the above network, each database is separated into two parts randomly in the experiments, with equal number of images as training (atlas) and testing (target) data sets. For the atlas pixels within region of interest (ROI), their surrounding patches (with the size of $20\times 20$ pixels) are extracted as training data, together with their corresponding labels. With the well trained network, we can obtain the structural signature for each pixel, by using its surrounding patch as input and extracting the feature vector before the output layer. 
\begin{table}
\centering
\caption{Detailed parameter settings of CNN network to extract structural signature.}
\includegraphics[width=\linewidth]{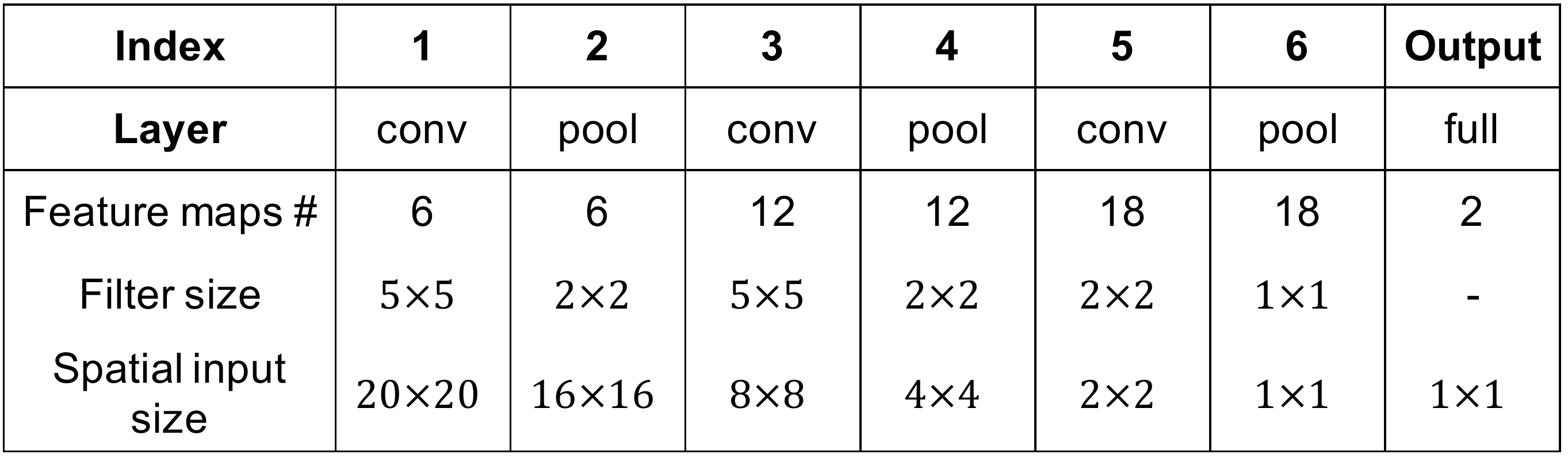}
\label{tab1}
\end{table}
\begin{table}
\centering
\caption{Summary of generated features for each pixel.}
\includegraphics[width=0.85\linewidth]{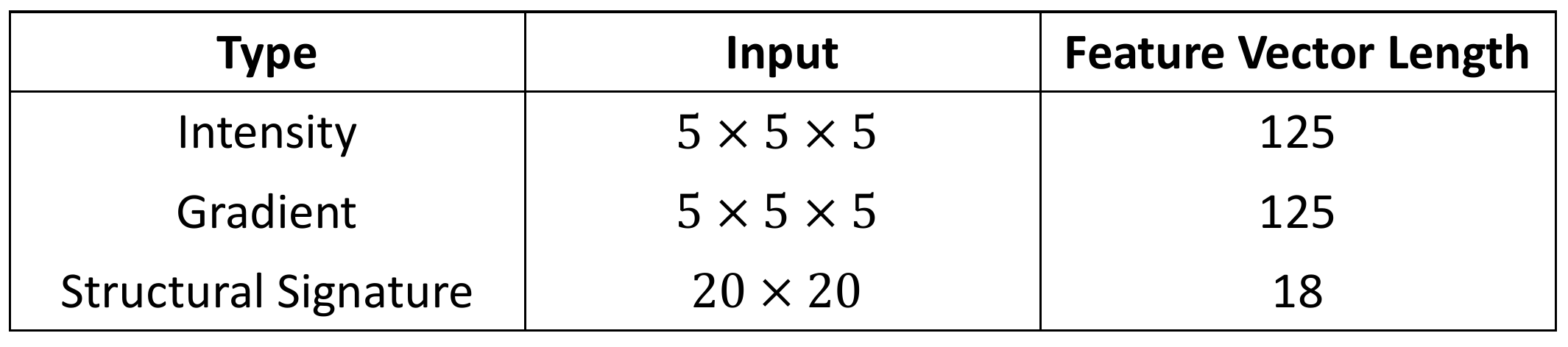}
\label{tab2}
\end{table}

As demonstrated in \cite{wang2014mitosis}, the performance of mitosis detection can be further improved by combining the discriminative CNN features with conventional handcrafted features, like morphology or color information. The method \cite{bai2015multi} also suggests that embracing high-level and low-level features yields better results in label fusion for cardiac image segmentation. Under the assumption that distinct features can assist the segmentation in a complementary way, in this paper, we extract multiple features from brain MR images and will consider the feature sensitivity during label fusion. For each pixel, besides the structural signature, the intensity values and gradient magnitudes in the surrounding cube are also assembled as feature vectors. In total, three kinds of feature vectors are generated in the proposed method, which is summarized in Table \ref{tab2}. 

\begin{figure}
\centering
\includegraphics[width=0.8\linewidth]{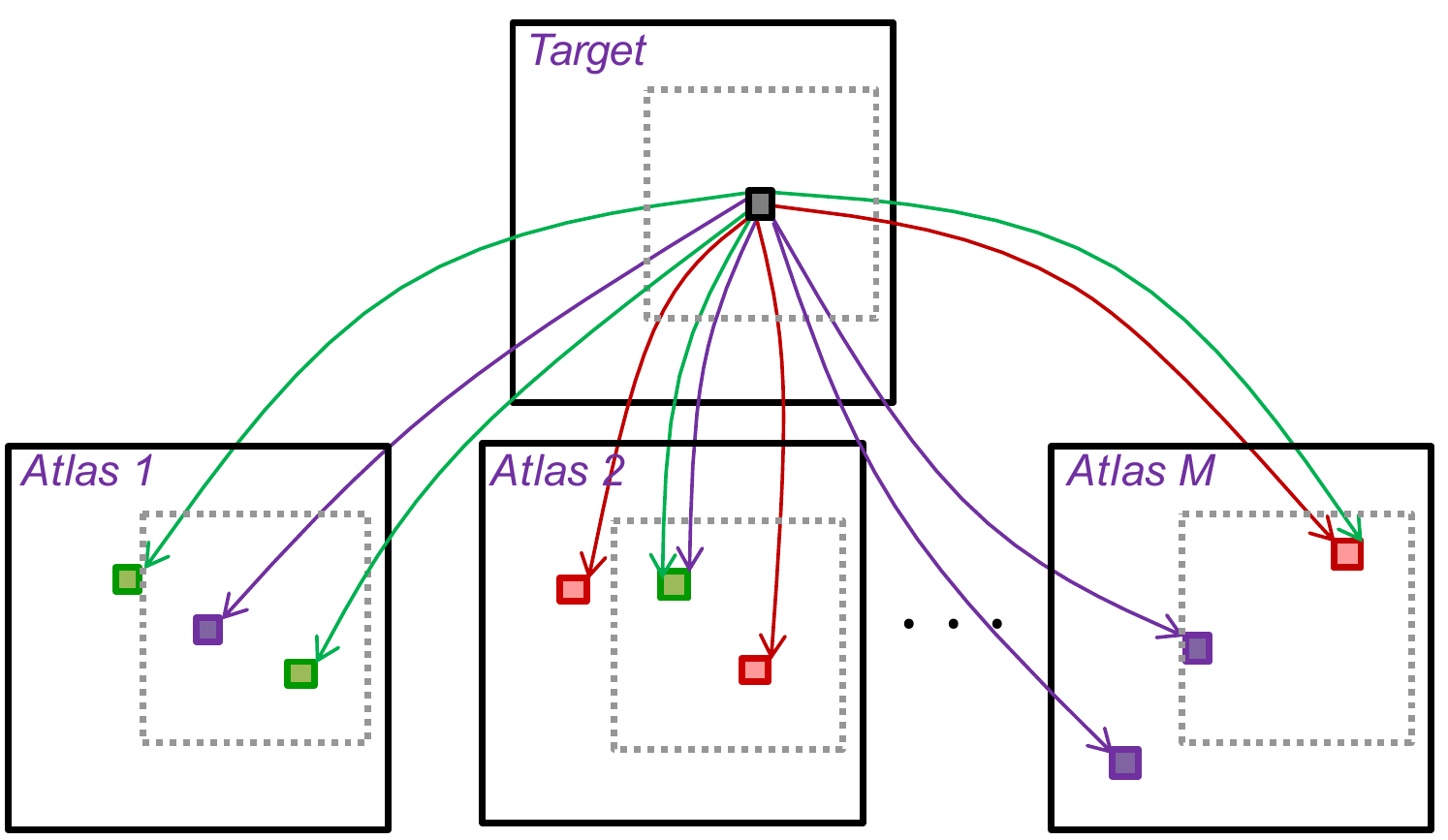}
\caption{Illustration of Feature Matching. For a considered pixel (Gray Square) in the target image, similar pixels are searched from the atlases using each kind of feature vector. Red, Purple and Green Squares represent similar pixels found with intensity, gradient and structural signature respectively. Dashed Gray Square is the spatial constraint and those outside similar pixels will not be involved as candidate nodes.}
\label{fig6}
\end{figure}
\subsection{Feature Matching}\label{FeaM}
After feature generation, the second stage in label fusion is to select candidate nodes from atlases. For each pixel in the target image, similar pixels can be selected from atlases using each kind of feature vector. In fact, it is the nearest neighbor (NN) problem to find similar points in real $d$-dimensional space from $N$ samples. As shown in Table \ref{tab2}, the dimension $d$ of our generated features (intensity, gradient and structural signature) has a value of 125, 125 and 18 respectively. Using the brute force approach to check each sample in a sequential order, the computation complexity can be $O(dN^2)$. Given the expensive computational cost of exact searching, approximate nearest neighbor (ANN) has been introduced to accelerate the searching speed. In \cite{arya1998optimal}, the ($1+\epsilon$)-approximation to $k$ nearest neighbors can be obtained in $O(kd\log N)$ time. However, the performance of this algorithm degrades rapidly along with the dimension increase and it cannot be applied well to high dimensional data. The matching results can become patchy when $d$ becomes as high as 20. To tackle this issue, several advanced ANN approaches have been proposed for the application on high dimensional data, such as randomized k-d tree \cite{silpa2008optimised}, locality sensitive hashing \cite{andoni2006near} and so on. In \cite{muja2014scalable}, it demonstrates that randomized k-d tree and priority search k-means tree \cite{muja2009fast} can obtain the best results through comprehensive experiments. Therefore, the feature matching component of the proposed method is carried out on the foundation of the randomized k-d tree provided by fast library for approximate nearest neighbors (FLANN) toolbox \cite{muja2014scalable}.

As shown in Fig. \ref{fig6}, similar pixels are selected from the atlases using randomized k-d tree with each kind of feature vector. Considering the poor contrast condition in MR brain images and similar histogram profiles among adjacent tissues, the atlas pixels selected with randomized k-d tree can belong to other structures and mislead the subsequent fusion procedure. As such, a spatial constraint is enforced in the proposed method to filter out the pixels which are too far away from the considered target pixel. For those atlas pixels which cannot meet the spatial constraint (i.e., outside Dashed Gray Square), they tend to be deceptive similar pixels and therefore are not involved in the pool of candidate nodes. 

\subsection{Feature Sensitive Label Prior}\label{FeaSLP}
In this paper, a novel method named Feature Sensitive Label Prior (FSLP) is proposed to capture label prior from atlases by seeking for the optimal linear combination of atlas nodes to reconstruct the feature vector of the target pixel, as illustrated in Fig. \ref{fig3}(a). For each considered pixel from the target image, its three kinds of features are extracted and concatenated together to formulate one augmented vector $y$. Its similar pixels selected from the atlases with Feature Matching are assembled as dictionary $A$. Given that the confidence and significance of different features can vary considerably, the feature coefficient $a_i$ is introduced to balance their influences. The optimal weight to reconstruct $y$ with dictionary $A$ is stored in vector $\beta$. The formulation of FSLP is given as follows:
\begin{figure}
\centering
\includegraphics[width=\linewidth]{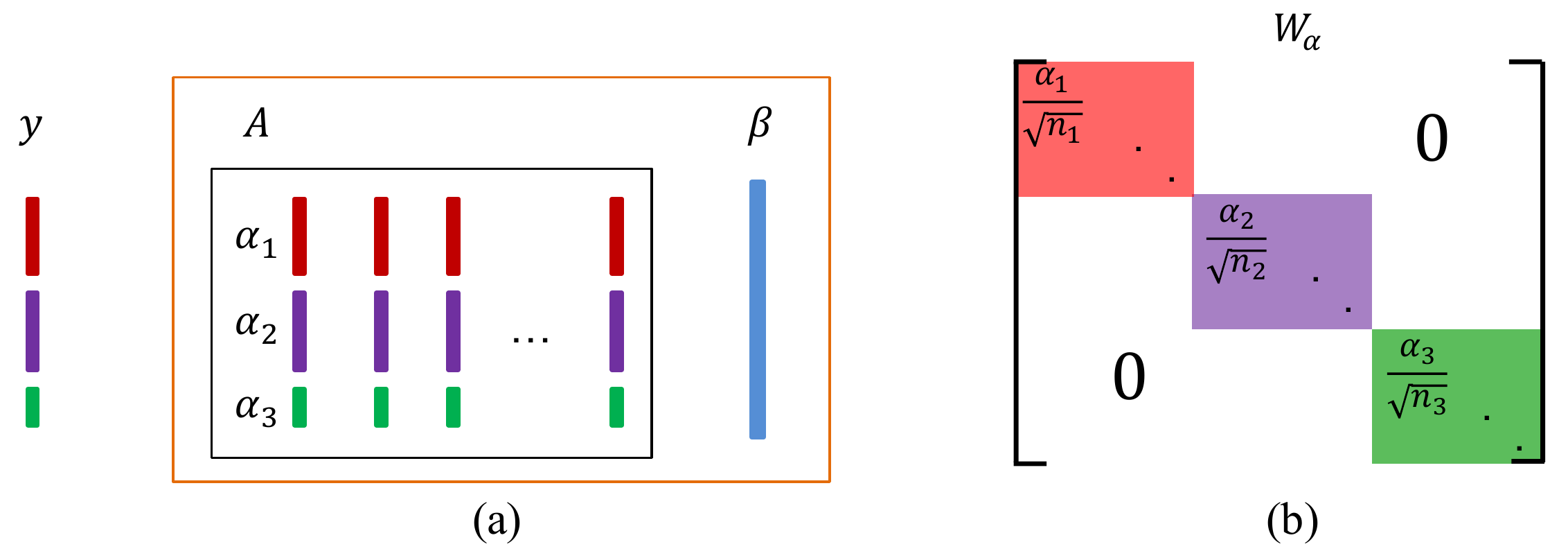}
\caption{\textbf{(a)} FSLP illustration. $y$ is the feature vector for the target pixel, concatenating intensity (Red), gradient (Purple) and structural signature (Green). $\alpha_i$ is feature coefficient and $A$ is a dictionary constructed with atlas feature vectors. $\beta$ is the vector storing reconstruction weights. \textbf{(b)} Feature sensitive matrix $W_\alpha$. Each diagonal sub-matrix (Red $W_1$, Purple $W_2$, Green $W_3$) corresponds to one kind of feature vector.}
\label{fig3}
\end{figure}
\begin{equation}\label{fslp}
\begin{split}
\min_{\alpha, \beta}~~& \frac{1}{2}|W_\alpha (y-A\beta)|_2^2+\lambda |\alpha|_2^2,\\
s.t.~~ & \sum_i \alpha_i=1,~\alpha_i\geq 0.
\end{split}
\end{equation}
$W_\alpha$ is the feature sensitive matrix, with its definition illustrated in Fig. \ref{fig3}(b). $W_\alpha$ is split into three subregions (Red $W_1$, Purple $W_2$ and Green $W_3$), each corresponding to one kind of feature vector. The diagonal elements in the sub-matrix $W_j$ are defined as:
\begin{equation}\label{wa}
\forall w_{ii} \in W_j, w_{ii}=\frac{\alpha_j}{\sqrt{n_j}}, 
\end{equation}
where $n_j$ is the length of the j-th feature vector. Through the division between the coefficient $\alpha_j$ and $\sqrt{n_j}$, the normalization on various features is enforced in the feature sensitive matrix. In Equation \eqref{fslp}, with the regularization term on the coefficient vector $\alpha$, it guarantees that no feature dominates the whole optimization procedure. 

By solving Equation \eqref{fslp}, optimal feature coefficient $\alpha$ and reconstruction weight $\beta$ can be obtained and label prior can be then estimated with grouped reconstruction error. However, Equation \eqref{fslp} is one non-convex problem, which may have multiple local optima and can be difficult to solve. The details of the proof are given in the Appendix. To solve Equation \eqref{fslp} efficiently, we also propose one solution for it in the following. 
~\\

\noindent \textbf{Problem Solution}

As discussed above, when optimizing $\alpha$ and $\beta$ simultaneously, Equation \eqref{fslp} is not a convex problem. To solve this non-convex problem efficiently, one heuristic approach is proposed in this paper by seeking optimal solutions for $\alpha$ and $\beta$ alternately. The first step is to fix $\alpha$ and Equation \eqref{fslp} turns into one least square problem:
\begin{equation}\label{fsp_a}
\min_{\beta}~~ |W_\alpha (y-A\beta)|_2^2.
\end{equation}
This optimization problem is convex and its solution is $\hat{\beta}=(W_\alpha A)\backslash(W_\alpha y)$. With updated $\beta$, the second step is to fix it and Equation \eqref{fslp} is then simplified to one quadratic programming problem:
\begin{equation}\label{fsp_b}
\min_{\alpha}~~ \frac{1}{2}\alpha^T \Lambda \alpha,~~~~s.t. \sum_i \alpha_i=1,~\alpha_i\geq 0,
\end{equation}
where \begin{center}
$\Lambda=\left[\begin{array}{lll}
\frac{\sum f_{1j}^2}{n_1}+\lambda & ~~~~~0 & ~~~~~0\\
~~~~~0 & \frac{\sum f_{2j}^2}{n_2}+\lambda & ~~~~~0\\
~~~~~0 & ~~~~~0 & \frac{\sum f_{3j}^2}{n_3}+\lambda
\end{array}
\right],$
\end{center}
\begin{center}
$f=y-A\beta=\left[\begin{array}{l}
f_1\\f_2\\f_3
\end{array}
\right].$
\end{center}
The newly introduced variables $f_1$, $f_2$ and $f_3$ are vectors related to three kinds of features, with length of $n_1$, $n_2$ and $n_3$ respectively. Equation \eqref{fsp_b} is also a convex problem and can be solved efficiently. The proposed heuristic algorithm iterates the above two steps until either one of the following two conditions are met: the change of $\alpha$ is below a threshold or iterations exceed the predefined number.

With $\alpha$ and $\beta$ acquired, the reconstruction error using each class can be estimated as follows:
\begin{equation}\label{rec}
e_F=|W_\alpha (y- A\beta_F)|^2,~e_B=|W_\alpha (y- A\beta_B)|^2, 
\end{equation}
where $F$ and $B$ refers to the foreground and background respectively. $\beta_F$ and $\beta_B$ refers to the weights for the foreground and background atlas nodes respectively. With the estimated reconstruction error, FSLP is encoded as edge weight on the graph during label fusion, which will be explained in next subsection.

\subsection{Label Fusion with Random Walker}\label{labelrw}
Besides the FSLP gathered from atlases, the anatomical knowledge from target images is also encoded in the proposed label fusion method. As mentioned in the Introduction section, to label pixels which locate deep inside or outside a subcortical structure is relatively easier as compared with those around structural boundary. Thanks to the location advantage, even with a rough initial label map generated by affine transformation, the labels of these pixels (far away from the object boundary) can be treated as confident results. This kind of confidence can be propagated to less confident pixels (near boundary) through image lattice, which is regarded as anatomical prior in our method.

In this paper, label fusion is formulated on an undirected graph $G=(V,E)$, where $V$ refers to a set of nodes consisting of foreground seeds $V_F$, background seeds $V_B$ and candidate nodes $V_C$. As both label and anatomical priors are employed in the proposed framework, two kinds of foreground seeds are included in $V_F$: $V_{F_a}$ from atlases and $V_{F_T}$ from the target image, similarly for $V_B$. As for $V_C$, it represents the set of nodes whose labels need to be determined during label fusion and these candidate nodes are selected from the target image. $E\subseteq V\times V$ is the set of edges $e_{ij}$ connecting nodes $v_i$ and $v_j$, with $w_{ij}$ as edge weight.

\begin{figure}
\centering
\includegraphics[width=\linewidth]{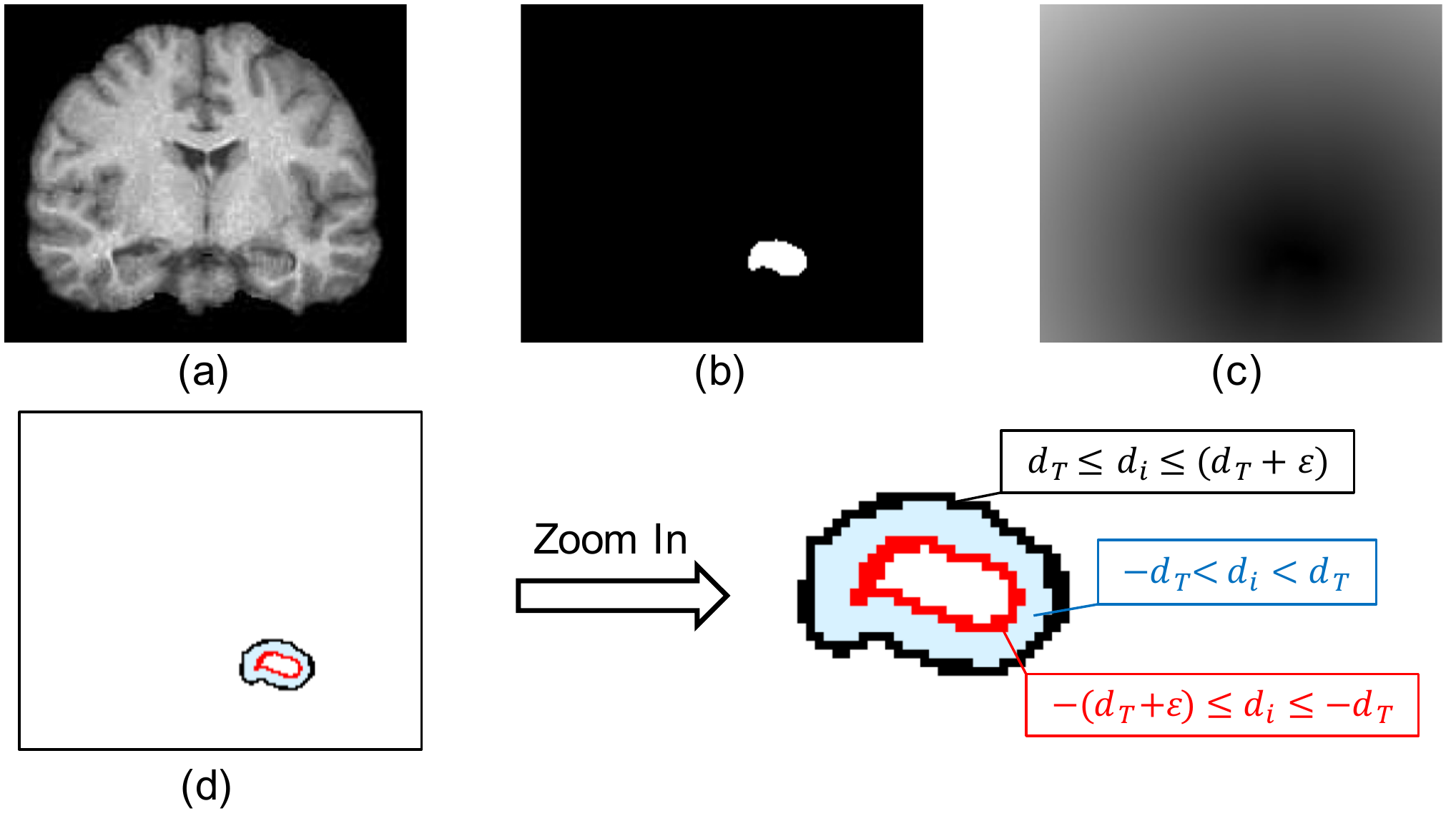}
\caption{Node selection. \textbf{(a)} 2D slice of target intensity image; \textbf{(b)} Initial label map fused with majority voting; \textbf{(c)} Signed Distance Map of \textbf{b}; \textbf{(d)} Red (inner) layer: target foreground seeds $-(d_T+\varepsilon)\leq d_i \leq -d_T$; Black (outer) layer: target background seeds $d_T\leq d_i \leq (d_T+\varepsilon)$; Blue (middle) layer: candidate nodes $-d_T< d_i < d_T$.}
\label{fig8}
\end{figure}

Since the number and location of nodes are critical to the efficiency of segmentation algorithms, the strategy for node selection needs to be deployed carefully. As the prior from atlases has been encoded to FSLP, $V_{F_a}$ and $V_{B_a}$ can be represented with two terminal nodes and the consideration of node selection can be limited to the target image. As discussed in our previous work \cite{bao2014label}, segmentation errors mainly lie around structural boundaries and those pixels which are far from the border can have higher label confidences. As such, in this paper, node selection is performed based on the Signed Distance Map (SDM), as illustrated in Fig. \ref{fig8}. With multiple label maps provided by a set of atlases, these maps are first fused with majority voting to produce the initial label map for the target image. Then its corresponding SDM can be estimated by calculating the Euclidean distance between a pixel and its nearest neighbour on the object boundary, with positive or negative value for outside (background) or inside (foreground) respectively. Using SDM and pre-defined distance threshold $d_T$, the target seeds and candidate nodes can be identified, as displayed in Fig. \ref{fig8}(d). 
\begin{figure}
\centering
\includegraphics[width=\linewidth]{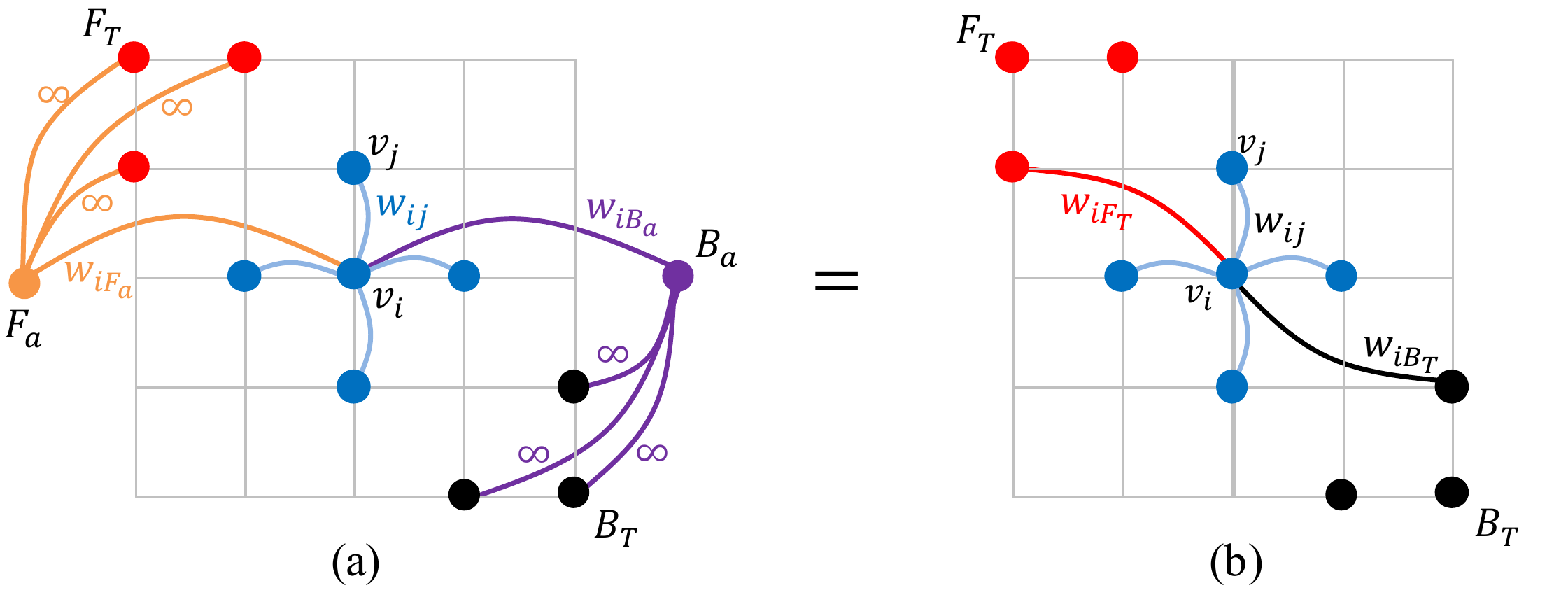}
\caption{Graph construction for label fusion. \textbf{(a)} Orange and Purple nodes: atlas seeds. Red and Black nodes: target foreground and background seeds. $v_i$ is one candidate node and $v_j$ is one of its neighbours, with $w_{ij}$ as edge weight. FSLP is encoded to $w_{iF_a}$ and $w_{iB_a}$. \textbf{(b)} An equal graph of \textbf{a}. }
\label{fig12}
\end{figure}

With seeds and candidate nodes settled, the graph for label fusion can be constructed with edge connections, as shown in Fig. \ref{fig12}(a). The Orange and Purple nodes represent the atlas seeds $V_{F_a}$ and $V_{B_a}$. Red and Black nodes refer to the foreground $V_{F_T}$ and background $V_{B_T}$ seeds selected from the target image. The influences of target seeds can be propagated to candidate nodes through image lattice. The affinity between nodes with lattice connection is defined using classical Gaussian function as follows:
\begin{equation}\label{wij}
\forall~v_j \in \mathcal{N}(v_i),~~~~w_{ij}=\exp(-\delta (I_T(v_i)-I_T(v_j))^2),
\end{equation}
where $v_i$ is one candidate node, $\mathcal{N}(v_i)$ refers to its 6-nearest neighbours in 3D image, $I_T(\cdot)$ is the pixel intensity value in the target image and $\delta$ is one tuning parameter. The FSLP is encoded as the edge weight between $v_i$ and atlas seeds, with the following definition:
\begin{equation}
w_{iF_a}=\frac{e_B}{e_F+e_B},~~~w_{iB_a}=\frac{e_F}{e_F+e_B}.
\end{equation}

Given that $V_{F_a}$ and $V_{F_T}$ are all foreground seeds, the edge weights between them are supposed to be infinity. In this case, setting up an edge between $v_i$ and $V_{F_a}$ is equal to appending an edge for $v_i$ with any target foreground seed, as illustrated in Fig. \ref{fig12}. In other words, the function of the atlas seeds can be replaced and FSLP can be assigned to the edges of $w_{iF_T}$ and $w_{iB_T}$ instead. In this way, the graph for label fusion can be constructed only with target nodes and the graph complexity can be greatly reduced. 

For graph-based image segmentation, the general energy function \cite{lezoray2012image} can be defined as follows:
\begin{equation}\label{gen}
\begin{split}
& E(x) =E_{unary}(x)+E_{binary}(x),\\
& =\sum _{v_i}~(w_{iF}^q|x_i-1|^p+w_{iB}^q|x_i-0|^p)+\sum _{e_{ij}}~w_{ij}^q|x_i-x_j|^p,
\raisetag{2.4\baselineskip}
\end{split}
\end{equation} 
where $x_i$ stands for the probability that node $v_i$ belongs to the foreground, with $x_F=1$ and $x_B=0$. The first unary term considers the data fidelity of each node independently and the second binary term takes the impact between connected nodes into account. By minimizing the above energy function, the optimal solution for $x$ can be obtained and the label of each node can be updated accordingly: $L(v_i)=1$ if $x_i\geq 0.5$ and $L(v_i)=0$ otherwise. 

As pointed out in \cite{couprie2011power}, by assigning different values to $p$ and $q$, Equation \eqref{gen} can be adapted to several popular image segmentation models, including Graph Cuts, Random Walker, Power Watershed, and so on. However, as Graph Cuts prefers a surface with minimum energy, it can suffer from surface shrink \cite{vicente2008graph}. In brain MR images, as a result of poor contrast conditions around structural boundaries, the shrinkage problem can be more serious. With Power Watershed, due to the fact that edge weights dominate the optimization procedure ($q$ set to infinity), the generated boundary can be rough \cite{couprie2011power}. As such, to obtain a smooth and quality segmentation result, we choose to employ Random Walker (RW), with $p$ and $q$ set to $2$. Then the minimization problem discussed above can be reformulated as follows:
\begin{equation}\label{obj}
\begin{split}
\min_{x}~~& \sum _{v_i}~[w_{iF_T}^2(x_i-1)^2+w_{iB_T}^2x_i^2]+\sum _{e_{ij}}~w_{ij}^2(x_i-x_j)^2,\\
s.t.~~~ & x_{F_T}=1,~x_{B_T}=0.
\raisetag{1.2\baselineskip}
\end{split}
\end{equation}
This equation can be viewed as a discrete Dirichlet problem and solved by using the Laplace equation with Dirichlet conditions through Graph Analysis Toolbox \cite{grady2003graph}.

Considering RW is sensitive to seed positions \cite{Sinop2007}, the foreground and background seeds need to be chosen carefully. As mentioned above, one fundamental step for node selection is the initial label map, whose quality depends on the choice of registration methods, for example, non-rigid or affine transformation. To increase the robustness of the proposed label fusion approach to registration procedure, an iterative RW scheme is introduced to update the label map and gradually improve the quality of node selection. The experimental results shown in Fig. \ref{fig9} also demonstrate that the segmentation accuracy can benefit from this iterative strategy and tends to be stable after several iterations.

\begin{figure}
\centering
\includegraphics[width=\linewidth]{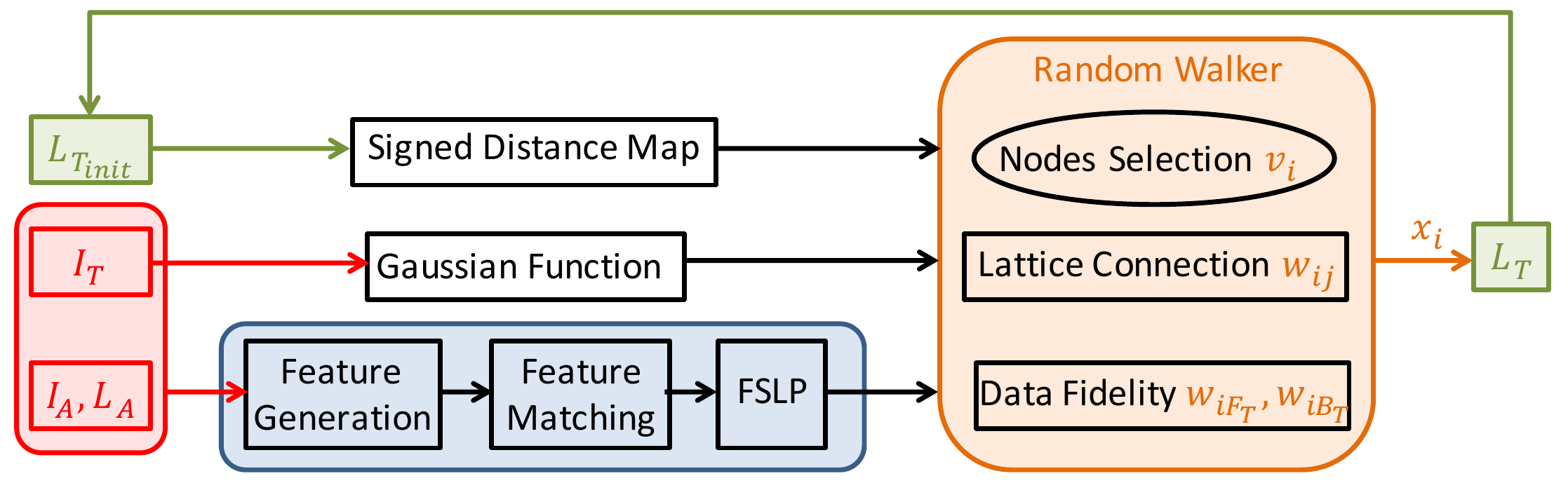}
\caption{Overview of the proposed feature sensitive label fusion.}
\label{fig7}
\end{figure}
The overview of the proposed method is summarized in Fig. \ref{fig7}. With atlas intensity $I_A$ and label maps $I_T$, affine transformation is first carried out and an initial label map for the target image $L_{T_{init}}$ is obtained with majority voting. Node selection can be performed based on the SDM of the initial label map and the graph for label fusion can be constructed with these target nodes. With intensity values, gradient magnitude and structural signature as augmented feature vector, candidate nodes are selected from atlases and the atlas prior is gathered in the form of FSLP. With the label prior from atlases and anatomical prior from the target itself, label fusion is formulated on a graph with Random Walker and the label map $L_T$ is updated gradually through iterations until stable.

\section{Experiments}
\subsection{Databases and Preprocessing}
To evaluate the performance of the proposed method, experiments have been carried out on two publicly available MR brain image databases -- IBSR\footnotemark[1] and LPBA40\footnotemark[2] \cite{shattuck2008construction}. The IBSR v2.0 database, consisting of 18 T1-weighted images with 84 manually labeled structures, is provided by the Center for Morphometries Analysis at Massachusetts General Hospital, U.S.A.. Three kinds of voxel resolutions ($mm^3$) are utilized in the IBSR database: $0.97\times 0.97\times 1.5$, $1.0\times 1.0\times 1.5$ and $0.84\times 0.84\times 1.5$. 18 healthy subjects, including 14 males and 4 females, took part in the image acquisition, with ages ranging between 7 and 71. All 18 images have been normalized to Talairach orientation and the bias field has been corrected.
\footnotetext[1]{http://www.nitrc.org/projects/ibsr}
\footnotetext[2]{http://www.loni.usc.edu/atlases/Atlas\_Detail.php?atlas\_id=12} 

The LPBA40 database, consisting of 40 images with 56 manually labeled structures and skull-stripped, is provided by the UCLA Laboratory of Neuro Imaging, U.S.A.. 40 human volunteers, including 20 males and 20 females, took part in the image acquisition, with ages ranging between 19 and 40. The 40 skull-stripped volumes have been rigidly registered to the MNI305 atlas and the intensity inhomogeneity has been corrected. Detailed description of these two databases is presented in Table \ref{dataset}.
\begin{table}
	\centering
	\caption{Description of IBSR and LPBA40 databases.}
	\includegraphics[width=\linewidth]{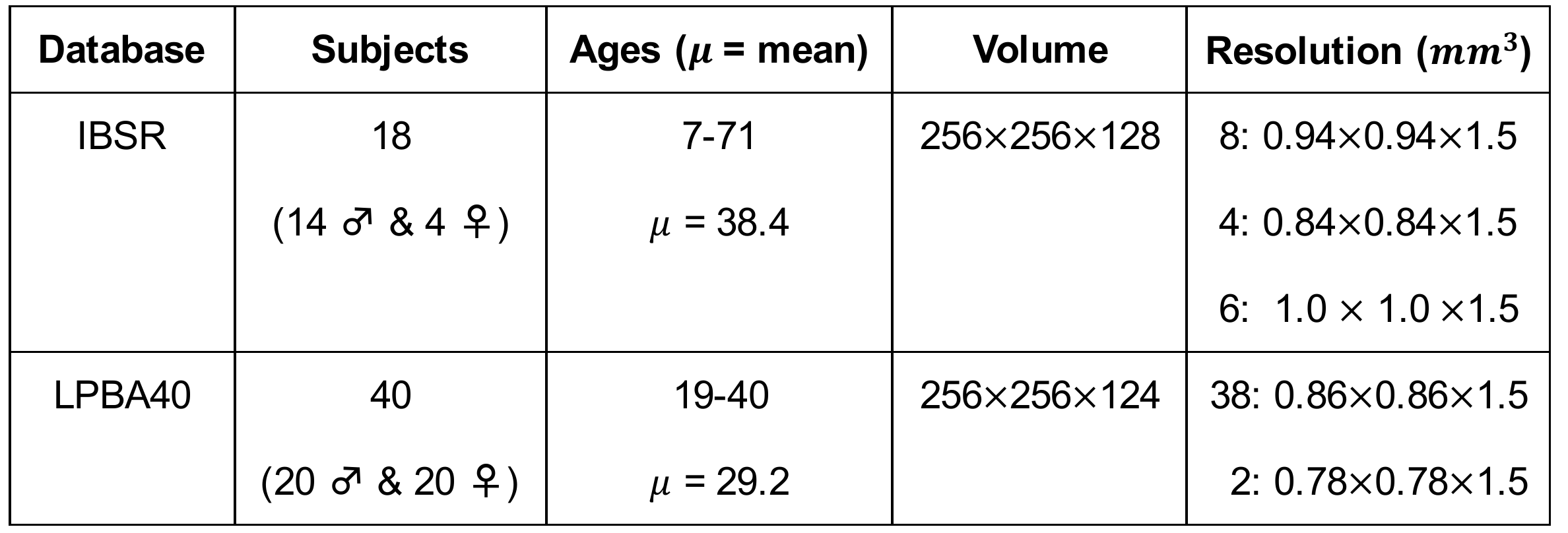}
	\label{dataset}
\end{table}

Given the significance of subcortical structures in clinical diagnosis, surgical planning and therapeutic assessment, in this paper, we focus on the extraction of subcortical structures from brain MR images. There are six subcortical structures labeled in IBSR database, including Amygdala, Caudate, Hippocampus, Pallidum, Putamen and Thalamus. As for the LPBA40 database, three subcortical structures are delineated: Caudate, Hippocampus and Putamen. Each of the subcortical structure has two sub-parts, located in the left and right hemispheres respectively. 

In the experiments, each database was separated into two parts randomly, with equal number of images as training (atlas) and testing (target) data sets. Considering the intensity inconsistency among images, histogram matching was first conducted with the Insight Toolkit\footnotemark[3].
\footnotetext[3]{http://www.itk.org/}
Then pair-wise registrations between each target image and all atlases were performed based on affine transformation, using FLIRT \cite{jenkinson2002improved} provided by FSL toolbox \cite{jenkinson2012fsl}. With multiple label maps generated with various atlases, majority voting was applied to generate the initial label map and the results were also employed as the baseline during comparison.

\subsection{Segmentation Results}
In the Methodology section, the label fusion method with Random Walker was initially designed for binary segmentation. While in the experiments, there are several remarkable subcortical structures in one brain volume and some of them can be adjacent with each other. Directly applying binary segmentation for each structure independently may cause some inconsistencies around the neighboring areas. As such, it is necessary to extend the binary segmentation to multi-class segmentation in a refined way. 

Distinct with other graph-based approaches (like Graph Cuts or Markov Random Field), Random Walker produces a probability map rather than a discrete label map, indicating the probability that each pixel belongs to the foreground. After applying Random Walker to each structure, we can obtain a vector $(p_{i1}, p_{i2},\cdots,p_{iK})$ for each pixel $v_i$, where $K$ is the total number of subcortical structures. $p_{ij}$ represents the probability that $v_i$ belongs to the $j$-th subcortical structure. As for the background probability, $p_{i0}$ is assigned as $1-\max(p_{i1}, p_{i2},\cdots,p_{iK})$. Then the probability distribution over the $K+1$ classes, including the background and multiple structures, can be estimated with the softmax function (normalized exponential function). The category with the largest probability is assigned as the final label for each pixel.

\begin{figure}
	\centering
	\includegraphics[width=\linewidth]{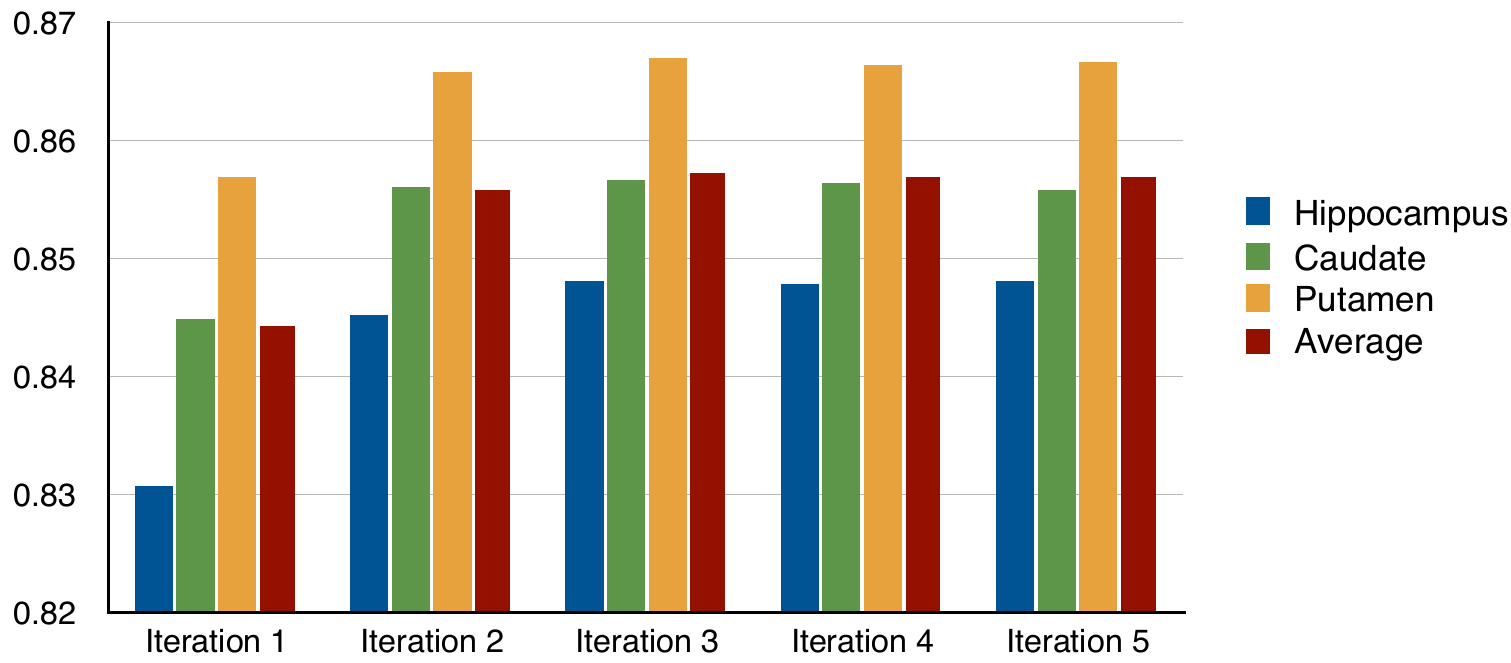}
	\caption{Segmentation results by our method at each iteration on LPBA40 database, measured with DC.}
	\label{fig9}
\end{figure}

Dice Coefficient (DC) is utilized to evaluate the quality of label fusion. In the proposed framework, the iterative strategy is exploited to update target label map $L_T$ gradually. To test the iterative effects, experiments have been conducted on LPBA40 database with available subcortical structures and the segmentation results at each iteration are recorded and displayed in Fig. \ref{fig9}. It can be observed that the segmentation accuracy increases with the number of iterations and tends to remain stable after three iterations.

\begin{figure}
	\centering
	\includegraphics[width=0.96\linewidth]{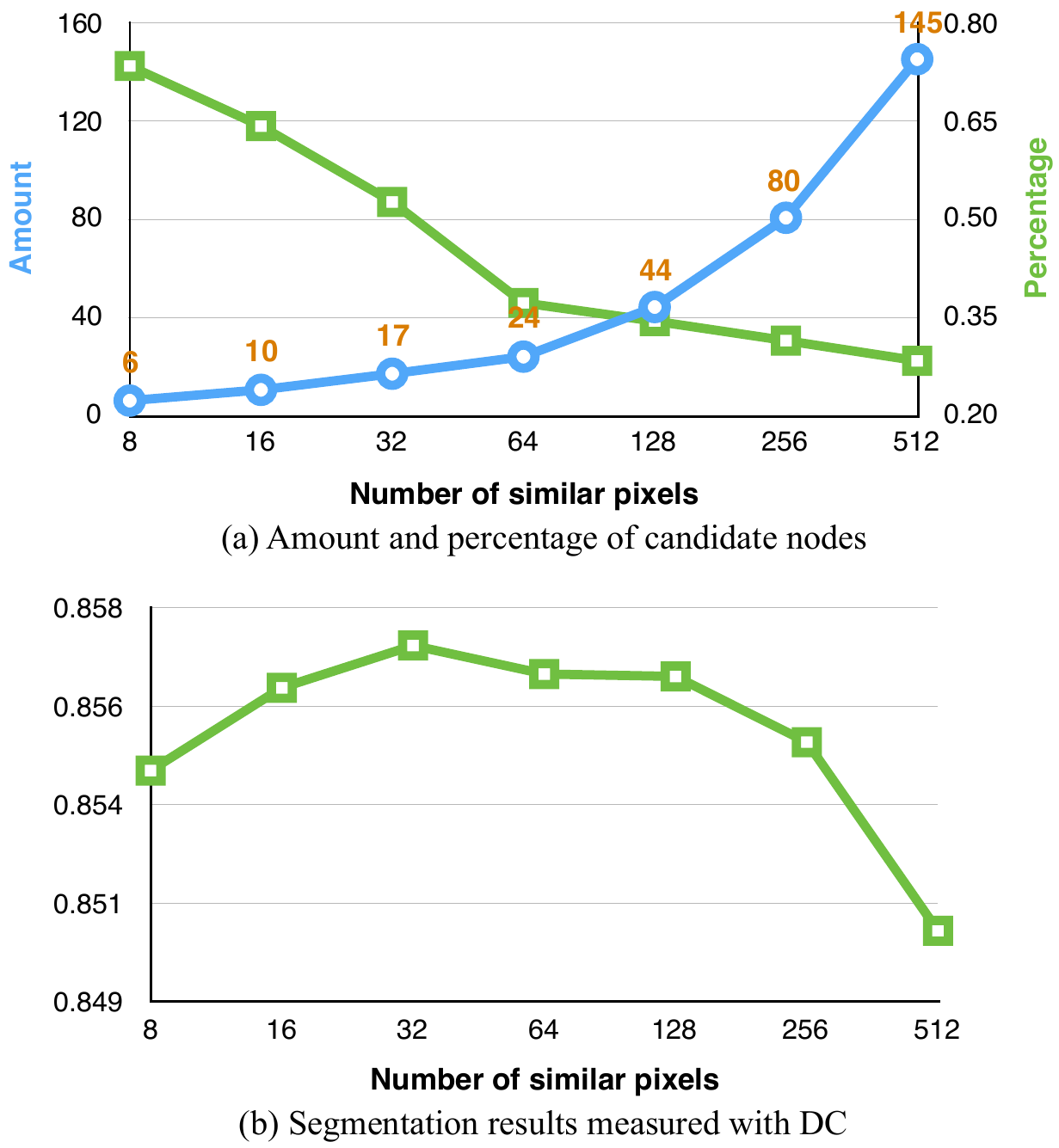}
	\caption{The effects of the setting of similar pixel amount on candidate nodes and segmentation quality with LPBA40 database. (a) Blue curve, amount of candidate nodes; Green curve, the percentage of candidate nodes among similar pixels. (b) Segmentation result measured with DC.}
	\label{fig10}
\end{figure}

Another set of experiments has been carried out to check how the amount of candidate nodes can influence the segmentation quality. As discussed in Feature Matching, with each kind of feature vector, a set of similar pixels can be collected from atlases with randomized k-d tree and the pool of candidate nodes can be further determined with a spatial constraint. In Fig. \ref{fig10}, the horizontal direction refers to the settings of how many similar pixels need to be selected with one kind of feature. In the Upper subfigure, the Bule curve displays the count of candidate nodes with three features, which indicates that with spatial constraint, only a portion of similar pixels can be kept in the candidate nodes pool. The Green curve shows the percentage of candidate nodes among similar pixels and the percentage decreases along with the increase of similar pixel amount, which can be caused by the disturbances from adjacent tissues with similar profiles. In the Bottom subfigure, the segmentation accuracy measured with DC is displayed and the peak of the performance lies around 32 similar pixels. Fig. \ref{fig10} demonstrates that segmentation quality is not proportional to the number of candidate nodes and the setting of 32 similar pixels gives the best performance. 
\begin{figure}
	\centering
	\includegraphics[width=0.85\linewidth]{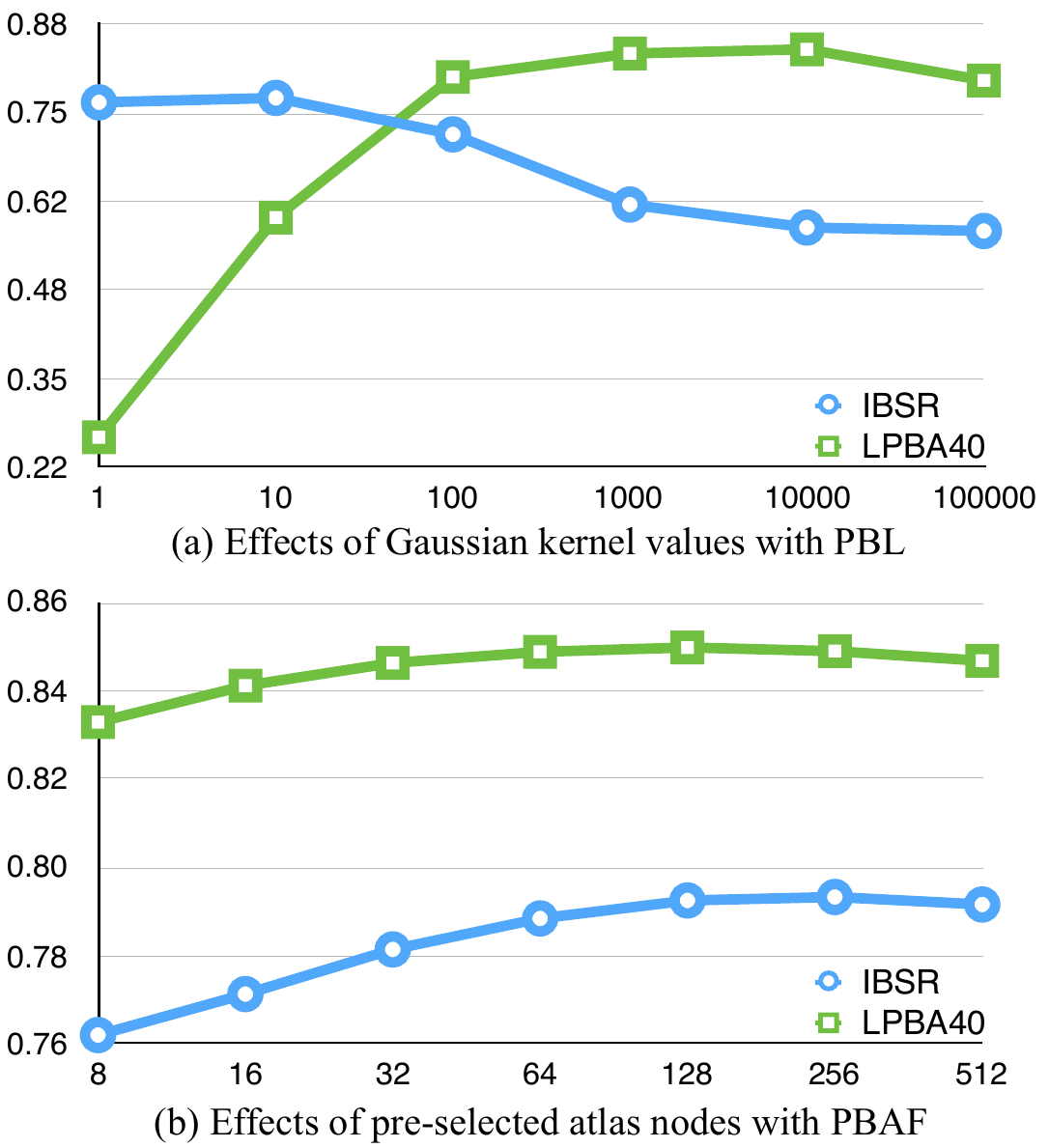}
	\caption{Parameters selection for compared methods, measured with DC.}
	\label{compare}
\end{figure}

Based on the preliminary test on LPBA40, for the experiments on IBSR, the iteration  was set to 3 and the number of similar atlas nodes was set to 32. The input patch size for various features follows Table \ref{tab2} and the spatial constraint during Feature Matching was set to $9\times 9\times 9$. In FSLP estimation, rather than choosing a fixed $\lambda$ value in Equation \eqref{fsp_b}, it was set to be adaptive $\lambda=\frac{1}{3}(\frac{\sum f_{1j}^2}{n_1}+\frac{\sum f_{2j}^2}{n_2}+\frac{\sum f_{3j}^2}{n_3})$ in each iteration. The settings of the rest parameters are listed as follows: signed distance threshold $d_T=2$ and $\varepsilon=1$ for node selection, and the tuning parameter used in Equation \eqref{wij} was set to $\delta=5$. 
\begin{table*}
	\caption{Experimental results on IBSR and LPBA databases, measured with DC. Highest values are written in Red.}
	\centering
	\includegraphics[width=0.98\linewidth]{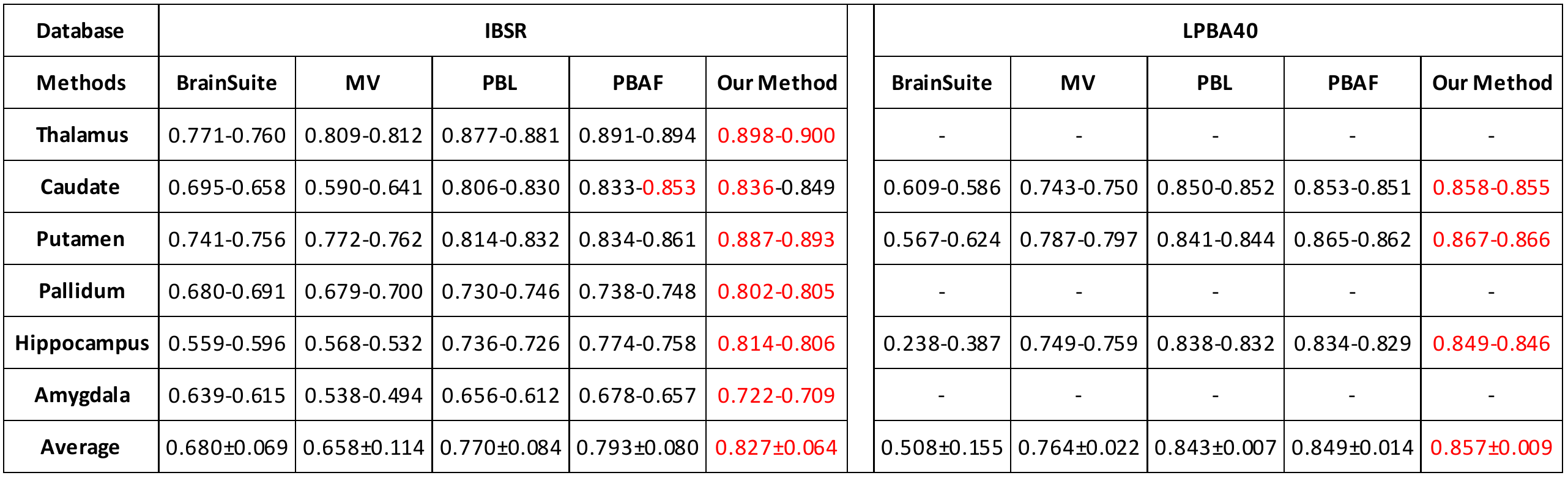}
	\label{reibsr}
\end{table*}
\begin{figure*}
	\centering
	\includegraphics[width=\linewidth]{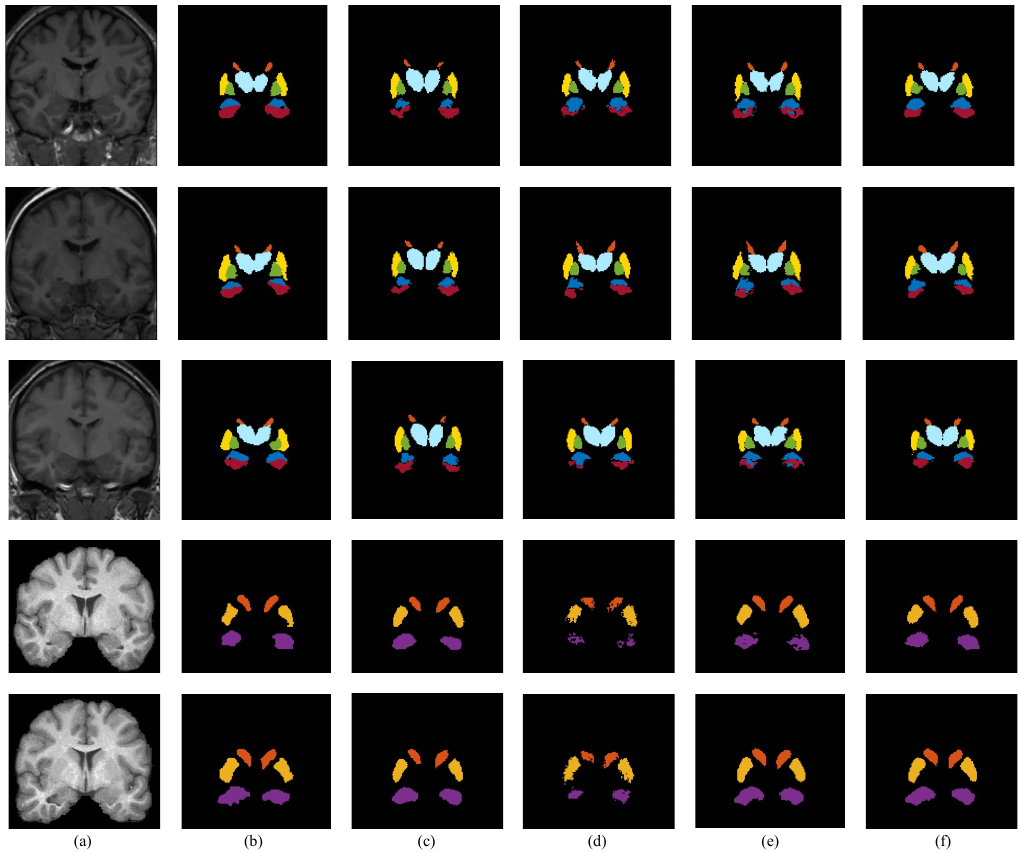}
	\caption{Visual segmentation results of 2D slices selected from 3D brain image volumes. Each row presents the original intensity slice, corresponding ground truth, the segmentation results generated with our method and other compared methods. The figures in the upper 3 rows are selected from IBSR database and those in the bottom 2 rows are from LBPA40 database. (a) 2D intensity slices from brain MR images; (b) Ground truth for reference; (c) Majority voting; (d) Patch-based label fusion (PBL); (e) Patch-based label fusion with augmented features (PBAF); (f) Our method.}
	\label{revis}
\end{figure*}

There are several existing softwares which support the automatic segmentation function for brain MR images, for example, BrainSuite \cite{shattuck2002brainsuite} or FreeSurfer \cite{fischl2012freesurfer}. Therefore, we decided to utilize BrainSuite, one of the available softwares, to label images in the IBSR and LPBA40 databases as a reference during evaluation. BrainSuite first runs surface/volume registration using the high-resolution ($0.5mm\times 0.5mm \times 0.8mm$) BCI-DNI\_brain atlas and then warps the label map from the atlas to the target image. Besides the reference BrainSuite and the baseline majority voting (MV), the comparison with the conventional patch-based label fusion (PBL) \cite{coupe2011patch} has been made for evaluation. Considering multiple features employed in the proposed method, it was also compared with the state-of-the-art method -- patch-based label fusion with augmented features (PBAF) \cite{bai2015multi}. In addition to intensity information, the spatial and context features are also appended for label fusion in PBAF. The implementations of PBL and PBAF provided by \cite{bai2015multi} were used in the experiments. To keep consistent in the evaluation, the patch size for PBL and PBAF was set to $5\times 5\times 5$ and the size of search volume was set to $9\times 9\times 9$. For PBL, the key parameter is the Gaussian kernel value and it was tested from $\{1,10,100,1000,10000,100000\}$ on two databases, measured with DC. From Fig. \ref{compare}(a), it can be observed that $10$ and $10000$ gives the best performance on IBSR and LPBA40 respectively. For PBAF, the parameter setting of pre-selected atlas nodes amount was tested from $\{2,16,32,64,128,256,512\}$ on two databases. Fig. \ref{compare}(b) indicates that $128$ can obtain the best result and the accuracy starts to decrease a little after the peak. 

The quantitative segmentation results on two databases generated with our method and compared methods are listed in Table \ref{reibsr}, with highest DC values written in Red. For the six subcortical structures delineated in the IBSR database, the accuracy on the left and right subcortical structures are listed respectively, separated by hyphen. The segmentation quality with available subcortical structures on the LPBA40 database is also reported in this Table. Although BrainSuite utilizes a high-resolution atlas, those patch-based methods (PBL, PBAF and our method) which rely on the low-resolution atlases inside the database, obtain much better performances. When compared with the baseline MV, our approach can create the considerable increase of 16.7 \% and 9.3\% respectively on two databases. The proposed method can still outperform the preeminent label fusion method PBAF by 3.2\% and 0.8\%. 

In Fig. \ref{revis}, we also present some visual results of 2D slices selected from 3D brain MR image volumes. Each row shows the original intensity slice, its corresponding ground truth, the segmentation results generated with compared methods and our method. The figures in the upper 3 rows are selected from IBSR database and those in the bottom 2 rows are from LBPA40 database. The first column (a) displays the 2D intensity slices from brain MR images, with the ground truth shown in column (b) for reference. The segmentation results generated with MV, PBL, PBAF and our method are shown in column (c) to (f). It can be observed that our method can obtain better segmentation quality. As compared with MV, the shapes of labeled subcortical structures by our method are closer to the ground truth. For the labeling results of the patch-based methods PBL and PBAF, some structures have isolate segments and the structural boundaries are relatively rough as compared with those of our method.

\subsection{Further Discussion}
\begin{figure}
	\centering 
	\includegraphics[width=\linewidth]{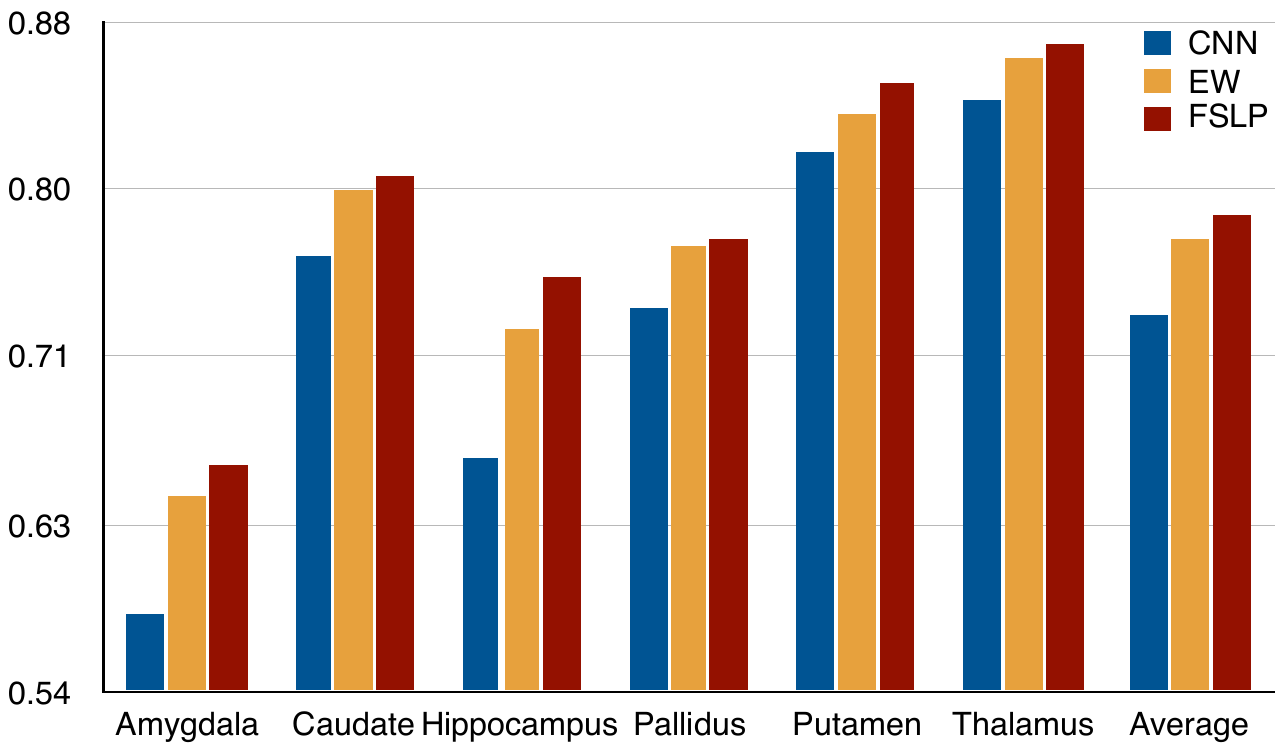}
	\caption{Comparison of the segmentation results on IBSR database using structural signature (extracted by CNN) alone, multiple features with equal weights (EW) and Feature Sensitive Label Prior (FSLP), measured with DC.}
	\label{ew_fslp}
\end{figure}
There is an underlying assumption for the proposed Feature Sensitive Label Prior (FSLP): distinct features can assist the segmentation in a complementary way. To test the effects of utilizing multiple features, we compare the preliminary FSLP results with the label fusion using structural signature alone, as displayed in Fig. \ref{ew_fslp}. In FSLP, besides the discriminative feature -- structural signature extracted by CNN, the less discriminative features -- intensity and gradient are also employed during label fusion. To further evaluate our feature sensitivity strategy, we also compare with the label fusion using fixed uniform feature coefficients, i.e., $\alpha_1=\alpha_2=\alpha_3$, and the results with equal weights (EW) are included in Fig. \ref{ew_fslp}. These results demonstrate that embracing distinct features can yield better performance and the feature sensitivity strategy can consistently improve the segmentation quality. It is noting that FSLP is a general method to capture label prior from multiple features and its usage is not limited to the three kinds of features. Other features, such as Local Binary Pattern (LBP) \cite{ojala2002multiresolution} or Histogram of Oriented Gradients (HoG) \cite{dalal2005histograms}, can be also encoded in FSLP to improve the performance. 

For the computation cost of the proposed method, there are four main components to be considered: feature generation, feature matching, FSLP and label fusion with Random Walker. During feature generation, three kinds of features are generated: intensity, gradient and structural signature. The complexity of intensity and gradient extraction is $O(dN)$, where $d$ is the feature length. As for the structural signature, it only needs one forward pass through the CNN network to obtain the feature vector. As discussed in Section \ref{FeaM}, the feature matching process is carried out with the efficient randomized k-d tree algorithm. The FSLP is a non-convex problem and one heuristic approach is designed for it by alternately solving two convex problems. Given that the value of objective function will decrease strictly during each iteration and this non-singular function is lower-bounded by a finite value, the heuristic approach will converge after several iterations \cite{kushner2012stochastic}. As both two convex problems (least square and quadratic programming) can be solved efficiently, the process to estimate FSLP can be finished in a short time. The last step is the label fusion with Random Walker, which is a discrete Dirichlet problem and can be solved efficiently using the Laplacian equation with the Dirichlet conditions. In total, the running time for labeling one sub-cortical structure in one target image is around 1.5 minutes using the proposed label fusion method (on a 3.1GHz, Quad-Core CPU with 8GB RAM machine), as compared with 10 minutes using PBAF.

\begin{figure}
	\centering
	\includegraphics[width=\linewidth]{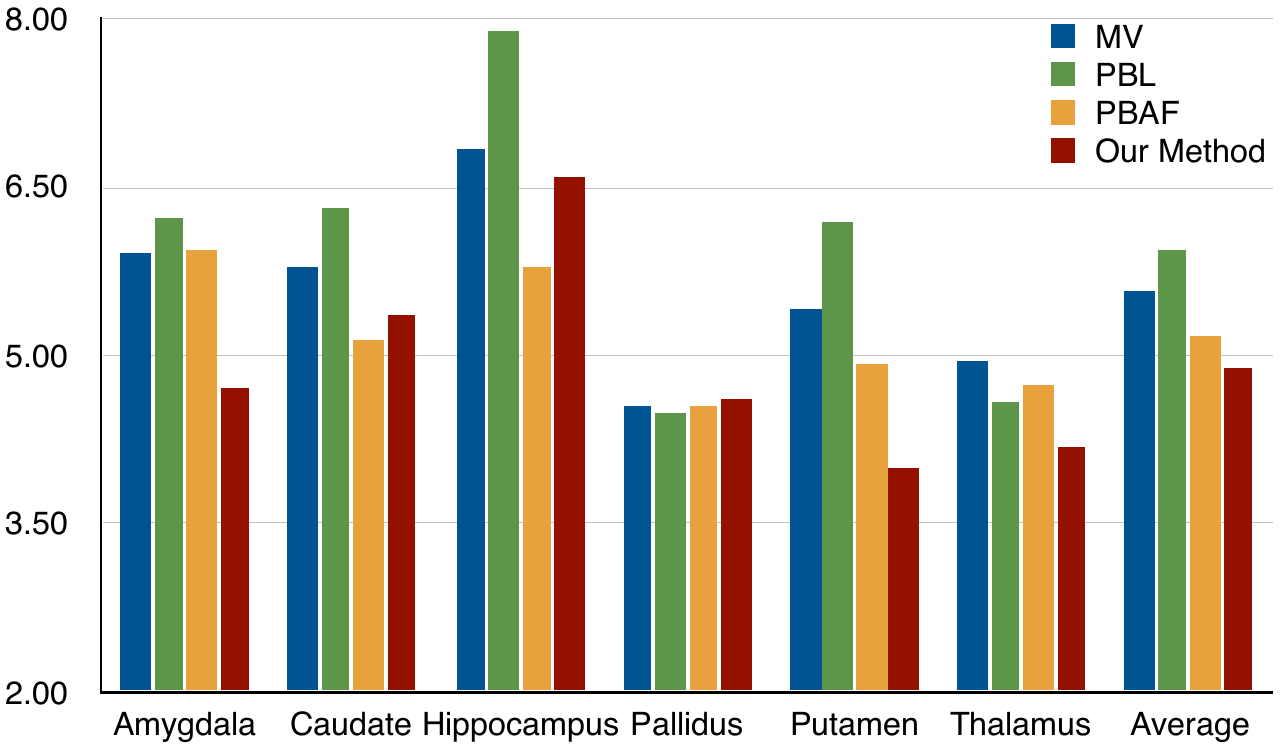}
	\caption{Segmentation quality on IBSR database, measured with HD.}
	\label{ibsrhd}
\end{figure}

Besides the volume overlap measurement Dice Coefficient (DC), we also evaluate the segmentation quality on IBSR database with one distance measurement -- Hausdorff Distance (HD). The segmentation results of six subcortical structures are shown in Fig. \ref{ibsrhd}, measured with HD. Although PBL can obtain higher DC values than MV, its performance is a little worse when measured with HD. This phenomenon may be caused by the lack of label consistency within the subcortical structures, as many holes and outliers exist in the labeled region (as shown in Fig. \ref{revis}(d)). The results measured with HD demonstrate that our method can still obtain competitive performance as compared with the state-of-the-art methods. 

\begin{figure}
	\centering 
	\includegraphics[width=\linewidth]{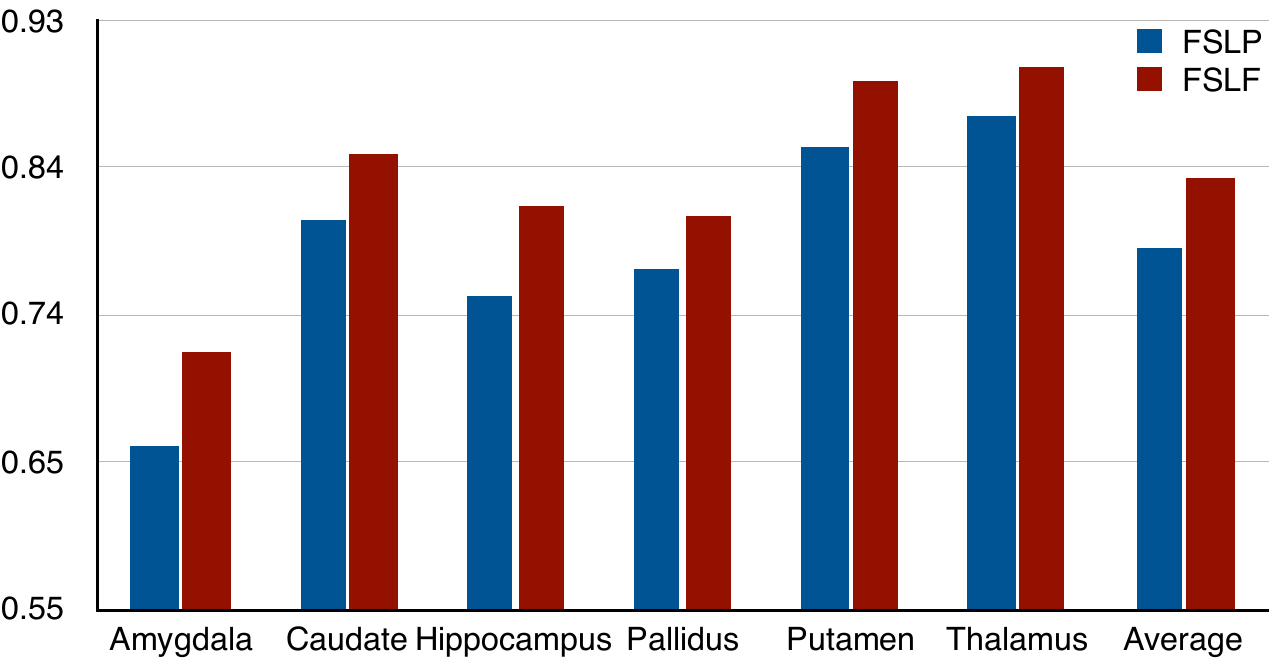}
	\caption{Comparison of the labeling result generated with FSLP and the complete proposed method on IBSR database, measured with DC.}
	\label{lref}
\end{figure}
\begin{figure*}
	\centering
	\includegraphics[width=0.9\linewidth]{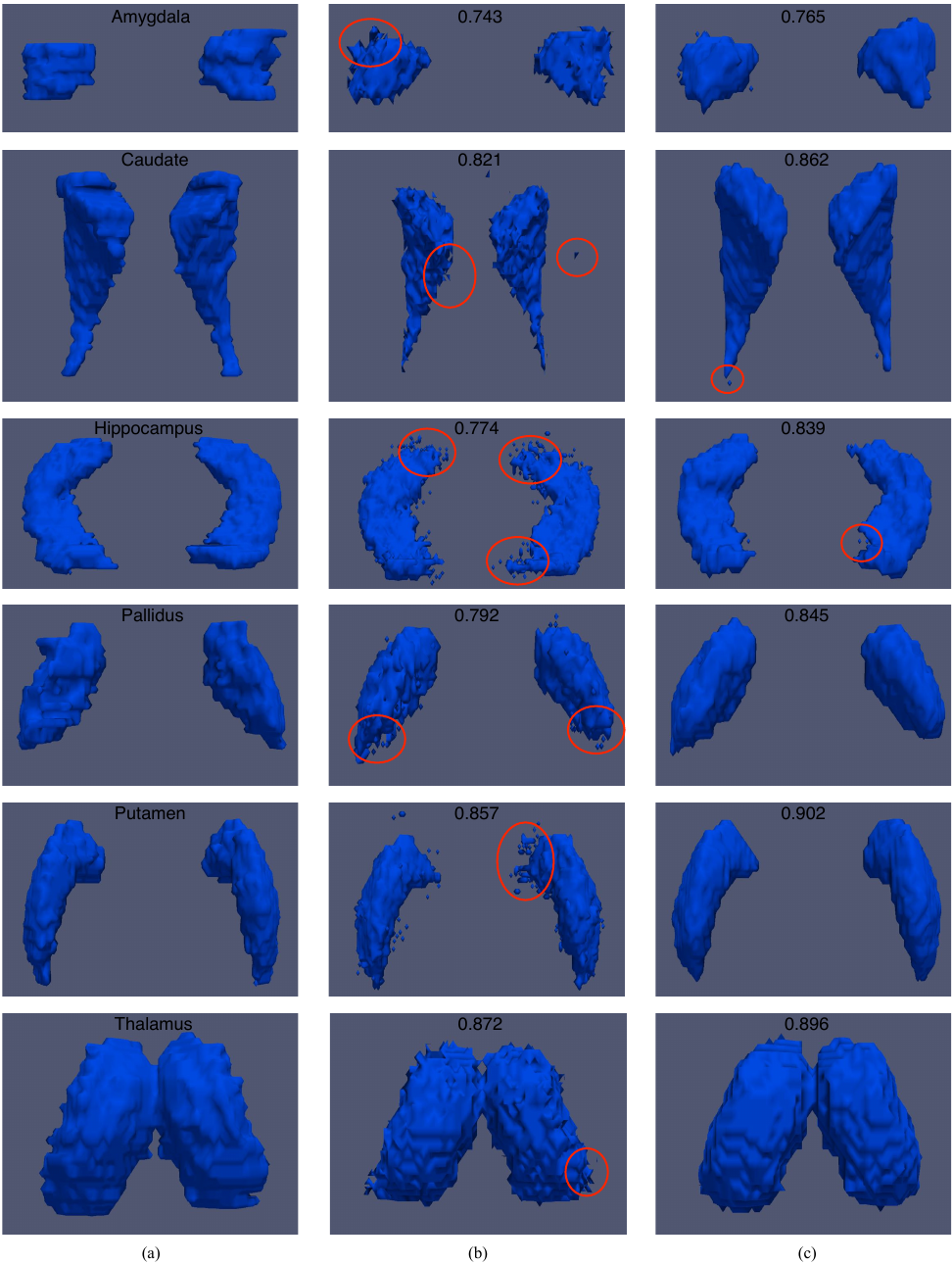}
	\caption{Some visual segmentation results of each subcortical structure on IBSR database. Each row presents the 3D labeled volumes for one selected subcortical structure. (a) Ground truth for reference; (b) Intermediate labeling result by FSLP; (c) Final segmentation after graph-based label fusion with anatomical priors.}
	\label{lrefv}
\end{figure*}

In the proposed method, we collect FSLP from atlases to capture the relationships between local intensity profiles and tissue labels, and utilize anatomical priors from target image to assist graph-based label fusion. To check the effects of two priors in detail, the preliminary segmentation with FSLP is estimated and compared with the result after label fusion. Based on Equation \eqref{rec}, the intermediate labeling result by FSLP can be generated by assigning labels to pixels with minimum reconstruction error. The comparison has been carried out on IBSR database, with quantitative results measured with DC. Fig. \ref{lref} indicates that embracing anatomical priors during label fusion can bring consistent improvements for the labeling of each subcortical structure. Some visual segmentation results for each subcortical structure on IBSR database are displayed in Fig. \ref{lrefv}. Each row presents the 3D labeled volumes provided by ground truth for reference, intermediate labeling by FSLP and the final segmentation after graph-based label fusion with anatomical priors. As shown in column (b), the labeling result by FSLP also suffers from the holes and outliers problems (Red Circles) as conventional patch-based methods. By introducing anatomical priors as graph seeds and lattice connections to enforce label consistency, although there are still some defects, the structural boundary becomes more smooth and the segmentation quality can be improved significantly as shown in column (c). The graph-based label fusion with Random Walker is an essential component in the proposed method, which can be also utilized to improve the labeling result by other conventional patch-based methods in the future.

\section{Conclusion}
In this paper, a novel framework for atlas-based image segmentation is proposed. It can effectively encode both the label priors from atlases and anatomical prior from the target image. Three kinds of features are employed to represent a pixel, including conventional intensity values and gradient magnitudes, together with the newly designed structural signature.


Besides the FSLP from atlases, the anatomical prior from the target itself is also employed for final label estimation. The label fusion process is formulated on a graph with Random Walker, with priors encoded as edge weights. Although atlas seeds involved in graph construction, an equal but simpler graph can be inferred which just relies on nodes from target image. The iterative strategy is employed to update the target label map gradually. The proposed framework has been compared with other state-of-the-art methods for comprehensive evaluation and experimental results indicate that it can obtain better label fusion quality consistently on two publicly available databases.  \\

\appendix
\noindent \textbf{Proof of Non-convexity}

To demonstrate the non-convexity of Equation \eqref{fslp}, one simple instance is first proven to be non-convex and then it can be derived that the original is also non-convex. Considering this simplified scenario -- the length of each feature vector is 1, the formulation of $W_\alpha$ becomes 
\begin{center}
$W_\alpha=\left[\begin{array}{lll}
\alpha_1 & 0 & 0\\
0 & \alpha_2 & 0\\
0 & 0 & \alpha_3
\end{array}
\right].$
\end{center}
By defining a new variable $f=y-A\beta=\left[\begin{array}{l}
f_1\\f_2\\f_3
\end{array}
\right]$, the format of the initial problem turns into: 
\begin{equation}\label{pro}
\begin{split}
\min~~& E=\alpha_1^2f_1^2+\alpha_2^2f_2^2+\alpha_3^2f_3^2+\lambda (\alpha_1^2+\alpha_2^2+\alpha_3^2),\\
s.t.~~ & \sum_i \alpha_i=1,~\alpha_i\geq 0.
\end{split}
\end{equation}
Let us consider a special case $f_1=f_2=f_3=f_*$ and use $\eta^2$ to represent $\alpha_1^2+\alpha_2^2+\alpha_3^2$ for simplification. Equation \eqref{pro} can be rewritten as follows:
\begin{equation}\label{relaxfsp}
\begin{split}
\min~~& E=\eta^2f_*^2+\lambda \eta^2,\\
s.t.~~ & 0 \leq \eta^2 \leq 1.
\end{split}
\end{equation}
Given the constraints $\sum_i \alpha_i=1$ and $\alpha_i\geq 0$, the range of $\eta^2$ becomes $0 \leq \eta^2 \leq 1$. If the special case shown in Equation \eqref{relaxfsp} is non-convex, it can be inferred that Equation \eqref{pro} is also non-convex. For this special case, it is much easier to determine whether it is convex or not. A problem is convex if and only if its Hessian matrix is positive semidefinite \cite{boyd2004convex}. The Hessian matrix of Equation \eqref{relaxfsp} can be calculated as follows:
\begin{equation}\label{hes}
H=\left[\begin{array}{ll}
~\frac{\partial^2E}{\partial \eta^2} & \frac{\partial^2E}{\partial \eta \partial f_*}\\
\frac{\partial^2E}{\partial f_* \partial \eta} & ~\frac{\partial^2E}{\partial f_*^2}
\end{array}
\right]
=\left[\begin{array}{ll}
2f_*^2+2\lambda & 4f_*\eta\\
~~4f_*\eta & ~2\eta^2
\end{array}
\right]
\end{equation}
If $H$ is positive semidefinite, all of its eigenvalues have to be non-negative. The determinant of $H-\gamma I$ is: 
\begin{center}
$det(H-\gamma I)=\gamma ^2-2(f_*^2+\eta^2+\lambda)\gamma+4\lambda \eta^2-12f_*^2\eta^2$.
\end{center}
The eigenvalues of $H$ are
\begin{center}
$\gamma=(f_*^2+\eta^2+\lambda)\pm \sqrt{(f_*^2+\eta^2+\lambda)^2-(4\lambda \eta^2-12f_*^2\eta^2)}$.
\end{center}
As it is not guaranteed that $\lambda > 3f_*^2$, one negative eigenvalue can appear. In this case, the Hessian matrix is not positive semidefinite and the special case shown in Equation \eqref{relaxfsp} is non-convex. It can be inferred that the original Equation \eqref{fslp} is also a non-convex problem.
~\\


\ifCLASSOPTIONcaptionsoff
  \newpage
\fi

\bibliographystyle{IEEEtran}
\bibliography{fslf_bib}

\begin{thebibliography}{10}
\providecommand{\url}[1]{#1}
\csname url@samestyle\endcsname
\providecommand{\newblock}{\relax}
\providecommand{\bibinfo}[2]{#2}
\providecommand{\BIBentrySTDinterwordspacing}{\spaceskip=0pt\relax}
\providecommand{\BIBentryALTinterwordstretchfactor}{4}
\providecommand{\BIBentryALTinterwordspacing}{\spaceskip=\fontdimen2\font plus
\BIBentryALTinterwordstretchfactor\fontdimen3\font minus
  \fontdimen4\font\relax}
\providecommand{\BIBforeignlanguage}[2]{{%
\expandafter\ifx\csname l@#1\endcsname\relax
\typeout{** WARNING: IEEEtran.bst: No hyphenation pattern has been}%
\typeout{** loaded for the language `#1'. Using the pattern for}%
\typeout{** the default language instead.}%
\else
\language=\csname l@#1\endcsname
\fi
#2}}
\providecommand{\BIBdecl}{\relax}
\BIBdecl

\bibitem{fischl2012freesurfer}
B.~Fischl, ``Freesurfer,'' \emph{Neuroimage}, vol.~62, no.~2, pp. 774--781,
  2012.

\bibitem{shattuck2002brainsuite}
D.~W. Shattuck and R.~M. Leahy, ``Brainsuite: an automated cortical surface
  identification tool,'' \emph{Medical image analysis}, vol.~6, no.~2, pp.
  129--142, 2002.

\bibitem{goebel2006analysis}
R.~Goebel, F.~Esposito, and E.~Formisano, ``Analysis of functional image
  analysis contest (fiac) data with brainvoyager qx: From single-subject to
  cortically aligned group general linear model analysis and self-organizing
  group independent component analysis,'' \emph{Human brain mapping}, vol.~27,
  no.~5, pp. 392--401, 2006.

\bibitem{geffroy2011brainvisa}
D.~Geffroy, D.~Rivi{\`e}re, I.~Denghien, N.~Souedet, S.~Laguitton, and
  Y.~Cointepas, ``Brainvisa: a complete software platform for neuroimaging,''
  in \emph{Python in neuroscience workshop}.\hskip 1em plus 0.5em minus
  0.4em\relax Euroscipy, Paris, 2011, pp. 15--16.

\bibitem{schnabel2001generic}
J.~A. Schnabel, D.~Rueckert, M.~Quist, J.~M. Blackall, A.~D. Castellano-Smith,
  T.~Hartkens, G.~P. Penney, W.~A. Hall, H.~Liu, C.~L. Truwit \emph{et~al.},
  ``A generic framework for non-rigid registration based on non-uniform
  multi-level free-form deformations,'' in \emph{Med. Image Comput.
  Computer-Assist. Intervent.}\hskip 1em plus 0.5em minus 0.4em\relax Springer,
  2001, pp. 573--581.

\bibitem{ashburner1999nonlinear}
J.~Ashburner and K.~J. Friston, ``Nonlinear spatial normalization using basis
  functions,'' \emph{Human Brain Mapping}, vol.~7, no.~4, pp. 254--266, 1999.

\bibitem{avants2008symmetric}
B.~B. Avants, C.~L. Epstein, M.~Grossman, and J.~C. Gee, ``Symmetric
  diffeomorphic image registration with cross-correlation: evaluating automated
  labeling of elderly and neurodegenerative brain,'' \emph{Med. Image Anal.},
  vol.~12, no.~1, pp. 26--41, 2008.

\bibitem{klein2009evaluation}
A.~Klein, J.~Andersson, B.~A. Ardekani, J.~Ashburner, B.~Avants, M.-C. Chiang,
  G.~E. Christensen, D.~L. Collins, J.~Gee, P.~Hellier \emph{et~al.},
  ``Evaluation of 14 nonlinear deformation algorithms applied to human brain
  mri registration,'' \emph{NeuroImage}, vol.~46, no.~3, pp. 786--802, 2009.

\bibitem{sabuncu2010generative}
M.~R. Sabuncu, B.~T. Yeo, K.~Van~Leemput, B.~Fischl, and P.~Golland, ``A
  generative model for image segmentation based on label fusion,'' \emph{IEEE
  Trans. Med. Imag.}, vol.~29, no.~10, pp. 1714--1729, 2010.

\bibitem{coupe2011patch}
P.~Coup{\'e}, J.~V. Manj{\'o}n, V.~Fonov, J.~Pruessner, M.~Robles, and D.~L.
  Collins, ``Patch-based segmentation using expert priors: Application to
  hippocampus and ventricle segmentation,'' \emph{NeuroImage}, vol.~54, no.~2,
  pp. 940--954, 2011.

\bibitem{liao2013sparse}
S.~Liao, Y.~Gao, J.~Lian, and D.~Shen, ``Sparse patch-based label propagation
  for accurate prostate localization in ct images,'' \emph{IEEE Trans. Med.
  Imag.}, vol.~32, no.~2, pp. 419--434, 2013.

\bibitem{rousseau2011supervised}
F.~Rousseau, P.~A. Habas, and C.~Studholme, ``A supervised patch-based approach
  for human brain labeling,'' \emph{IEEE Trans. Med. Imag.}, vol.~30, no.~10,
  pp. 1852--1862, 2011.

\bibitem{wang2013multi}
H.~Wang, J.~Suh, S.~Das, J.~Pluta, C.~Craige, and P.~Yushkevich, ``Multi-atlas
  segmentation with joint label fusion,'' \emph{IEEE Trans. Pattern Anal. Mach.
  Intell.}, vol.~35, no.~3, pp. 611--623, 2013.

\bibitem{tong2013segmentation}
T.~Tong, R.~Wolz, P.~Coup{\'e}, J.~V. Hajnal, D.~Rueckert, A.~D.~N. Initiative
  \emph{et~al.}, ``Segmentation of mr images via discriminative dictionary
  learning and sparse coding: Application to hippocampus labeling,''
  \emph{NeuroImage}, vol.~76, pp. 11--23, 2013.

\bibitem{ta2014optimized}
V.-T. Ta, R.~Giraud, D.~L. Collins, and P.~Coup{\'e}, ``Optimized patchmatch
  for near real time and accurate label fusion,'' in \emph{Med. Image Comput.
  Computer-Assist. Intervent.}, ser. LNCS, P.~Golland, N.~Hata, C.~Barillot,
  J.~Hornegger, and R.~Howe, Eds.\hskip 1em plus 0.5em minus 0.4em\relax
  Springer, 2014, vol. 8675, pp. 105--112.

\bibitem{coupe2008optimized}
P.~Coup{\'e}, P.~Yger, S.~Prima, P.~Hellier, C.~Kervrann, and C.~Barillot, ``An
  optimized blockwise nonlocal means denoising filter for 3-d magnetic
  resonance images,'' \emph{IEEE Trans. Med. Imag.}, vol.~27, no.~4, pp.
  425--441, 2008.

\bibitem{bai2015multi}
W.~Bai, W.~Shi, C.~Ledig, and D.~Rueckert, ``Multi-atlas segmentation with
  augmented features for cardiac mr images,'' \emph{Med. Image Anal.}, vol.~19,
  no.~1, pp. 98--109, 2015.

\bibitem{liu2009slep}
J.~Liu, S.~Ji, and J.~Ye, ``Slep: Sparse learning with efficient projections,''
  \emph{Arizona State University}, vol.~6, 2009.

\bibitem{koch2015multi}
L.~M. Koch, M.~Rajchl, T.~Tong, J.~Passerat-Palmbach, P.~Aljabar, and
  D.~Rueckert, ``Multi-atlas segmentation as a graph labelling problem:
  Application to partially annotated atlas data,'' in \emph{Information
  Processing in Medical Imaging}.\hskip 1em plus 0.5em minus 0.4em\relax
  Springer, 2015, pp. 221--232.

\bibitem{lecun1998gradient}
Y.~LeCun, L.~Bottou, Y.~Bengio, and P.~Haffner, ``Gradient-based learning
  applied to document recognition,'' \emph{Proceedings of the IEEE}, vol.~86,
  no.~11, pp. 2278--2324, 1998.

\bibitem{lowe2004distinctive}
D.~G. Lowe, ``Distinctive image features from scale-invariant keypoints,''
  \emph{International journal of computer vision}, vol.~60, no.~2, pp. 91--110,
  2004.

\bibitem{krizhevsky2012imagenet}
A.~Krizhevsky, I.~Sutskever, and G.~E. Hinton, ``Imagenet classification with
  deep convolutional neural networks,'' in \emph{Advances in Neural Information
  Processing Systems}, 2012, pp. 1097--1105.

\bibitem{long2015fully}
J.~Long, E.~Shelhamer, and T.~Darrell, ``Fully convolutional networks for
  semantic segmentation,'' in \emph{Proceedings of the IEEE Conference on
  Computer Vision and Pattern Recognition}, 2015, pp. 3431--3440.

\bibitem{sercu2015very}
T.~Sercu, C.~Puhrsch, B.~Kingsbury, and Y.~LeCun, ``Very deep multilingual
  convolutional neural networks for lvcsr,'' \emph{arXiv preprint
  arXiv:1509.08967}, 2015.

\bibitem{nair2010rectified}
V.~Nair and G.~E. Hinton, ``Rectified linear units improve restricted boltzmann
  machines,'' in \emph{International Conference on Machine Learning}, 2010, pp.
  807--814.

\bibitem{maas2013rectifier}
A.~L. Maas, A.~Y. Hannun, and A.~Y. Ng, ``Rectifier nonlinearities improve
  neural network acoustic models,'' in \emph{International Conference on
  Machine Learning}, vol.~30, 2013.

\bibitem{wang2014mitosis}
H.~Wang, A.~Cruz-Roa, A.~Basavanhally, H.~Gilmore, N.~Shih, M.~Feldman,
  J.~Tomaszewski, F.~Gonzalez, and A.~Madabhushi, ``Mitosis detection in breast
  cancer pathology images by combining handcrafted and convolutional neural
  network features,'' \emph{Journal of Medical Imaging}, vol.~1, no.~3, pp.
  034\,003--034\,003, 2014.

\bibitem{arya1998optimal}
S.~Arya, D.~M. Mount, N.~S. Netanyahu, R.~Silverman, and A.~Y. Wu, ``An optimal
  algorithm for approximate nearest neighbor searching fixed dimensions,''
  \emph{Journal of the ACM}, vol.~45, no.~6, pp. 891--923, 1998.

\bibitem{silpa2008optimised}
C.~Silpa-Anan and R.~Hartley, ``Optimised kd-trees for fast image descriptor
  matching,'' in \emph{Comput. Vis. Pattern Recognit.}\hskip 1em plus 0.5em
  minus 0.4em\relax IEEE, 2008, pp. 1--8.

\bibitem{andoni2006near}
A.~Andoni and P.~Indyk, ``Near-optimal hashing algorithms for approximate
  nearest neighbor in high dimensions,'' in \emph{IEEE Symposium on Foundations
  of Computer Science}.\hskip 1em plus 0.5em minus 0.4em\relax IEEE, 2006, pp.
  459--468.

\bibitem{muja2014scalable}
M.~Muja and D.~Lowe, ``Scalable nearest neighbour algorithms for high
  dimensional data,'' \emph{IEEE Trans. Pattern Anal. Mach. Intell.}, vol.~36,
  no.~11, pp. 2227--2240, 2014.

\bibitem{muja2009fast}
M.~Muja and D.~G. Lowe, ``Fast approximate nearest neighbors with automatic
  algorithm configuration.'' \emph{Proc. Int. Conf. Computer Vis. Theory
  Appl.}, vol.~2, pp. 331--340, 2009.

\bibitem{bao2014label}
S.~Bao and A.~C. Chung, ``Label inference with registration and patch priors,''
  in \emph{Med. Image Comput. Computer-Assist. Intervent.}, ser. LNCS,
  P.~Golland, N.~Hata, C.~Barillot, J.~Hornegger, and R.~Howe, Eds.\hskip 1em
  plus 0.5em minus 0.4em\relax Springer, 2014, vol. 8673, pp. 731--738.

\bibitem{lezoray2012image}
O.~L{\'e}zoray and L.~Grady, \emph{Image Processing and Analysis with Graphs:
  Theory and Practice}.\hskip 1em plus 0.5em minus 0.4em\relax CRC Press, 2012.

\bibitem{couprie2011power}
C.~Couprie, L.~Grady, L.~Najman, and H.~Talbot, ``Power watershed: A unifying
  graph-based optimization framework,'' \emph{IEEE Trans. Pattern Anal. Mach.
  Intell.}, vol.~33, no.~7, pp. 1384--1399, 2011.

\bibitem{vicente2008graph}
S.~Vicente, V.~Kolmogorov, and C.~Rother, ``Graph cut based image segmentation
  with connectivity priors,'' in \emph{Comput. Vis. Pattern Recognit.}, 2008,
  pp. 1--8.

\bibitem{grady2003graph}
L.~Grady and E.~Schwartz, ``The graph analysis toolbox: Image processing on
  arbitrary graphs,'' Tech. Rep. 021, 2003.

\bibitem{Sinop2007}
A.~Sinop and L.~Grady, ``A seeded image segmentation framework unifying graph
  cuts and random walker which yields a new algorithm,'' in \emph{International
  Conference on Computer Vision}, 2007, pp. 1--8.

\bibitem{shattuck2008construction}
D.~W. Shattuck, M.~Mirza, V.~Adisetiyo, C.~Hojatkashani, G.~Salamon, K.~L.
  Narr, R.~A. Poldrack, R.~M. Bilder, and A.~W. Toga, ``Construction of a 3d
  probabilistic atlas of human cortical structures,'' \emph{NeuroImage},
  vol.~39, no.~3, pp. 1064--1080, 2008.

\bibitem{jenkinson2002improved}
M.~Jenkinson, P.~Bannister, M.~Brady, and S.~Smith, ``Improved optimization for
  the robust and accurate linear registration and motion correction of brain
  images,'' \emph{NeuroImage}, vol.~17, no.~2, pp. 825--841, 2002.

\bibitem{jenkinson2012fsl}
M.~Jenkinson, C.~F. Beckmann, T.~E. Behrens, M.~W. Woolrich, and S.~M. Smith,
  ``Fsl,'' \emph{NeuroImage}, vol.~62, no.~2, pp. 782--790, 2012.

\bibitem{ojala2002multiresolution}
T.~Ojala, M.~Pietikainen, and T.~Maenpaa, ``Multiresolution gray-scale and
  rotation invariant texture classification with local binary patterns,''
  \emph{IEEE Trans. Pattern Anal. Mach. Intell.}, vol.~24, no.~7, pp. 971--987,
  2002.

\bibitem{dalal2005histograms}
N.~Dalal and B.~Triggs, ``Histograms of oriented gradients for human
  detection,'' in \emph{Comput. Vis. Pattern Recognit.}, vol.~1.\hskip 1em plus
  0.5em minus 0.4em\relax IEEE, 2005, pp. 886--893.

\bibitem{kushner2012stochastic}
H.~J. Kushner and D.~S. Clark, \emph{Stochastic approximation methods for
  constrained and unconstrained systems}.\hskip 1em plus 0.5em minus
  0.4em\relax Springer Science \& Business Media, 2012.

\bibitem{boyd2004convex}
S.~Boyd and L.~Vandenberghe, \emph{Convex optimization}.\hskip 1em plus 0.5em
  minus 0.4em\relax Cambridge university press, 2004.

\end{thebibliography}

\end{document}